\documentclass{article} 
\usepackage{times}
\usepackage[preprint]{neurips_2026}

\usepackage{amsmath,amsfonts,bm}

\def\eqref#1{equation~\ref{#1}}

\def\1{\bm{1}}

\DeclareMathAlphabet{\mathsfit}{\encodingdefault}{\sfdefault}{m}{sl}
\SetMathAlphabet{\mathsfit}{bold}{\encodingdefault}{\sfdefault}{bx}{n}

\newcommand{\R}{\mathbb{R}}

\usepackage{hyperref}
\usepackage{url}
\usepackage{graphicx}
\usepackage{caption}
\usepackage{subcaption}
\usepackage{booktabs}
\usepackage[shortlabels]{enumitem}
\usepackage{multirow}
\usepackage{arydshln}
\usepackage{rotating}

\usepackage{tikz}
\usetikzlibrary{positioning,arrows.meta}
\usetikzlibrary{calc}

\usepackage{cleveref}
\usepackage{autonum}
\usepackage{xspace}

\newcommand{\x}{x}

\newcommand{\axi}[1][i]{\x_{#1}}

\newcommand{\pseqxn}[1][n]{(\axi[i])_{i\geq 1}} 
\newcommand{\pseqxnn}[1][n]{(\axi[i])_{i=1}^n} 

\def\balign#1\ealign{\begin{align}#1\end{align}}
\def\baligns#1\ealigns{\begin{align*}#1\end{align*}}
\def\balignat#1\ealign{\begin{alignat}#1\end{alignat}}
\def\balignats#1\ealigns{\begin{alignat*}#1\end{alignat*}}
\def\bitemize#1\eitemize{\begin{itemize}#1\end{itemize}}
\def\benumerate#1\eenumerate{\begin{enumerate}#1\end{enumerate}}

\newenvironment{talign*}
 {\csname align*\endcsname}
 {\endalign}
\newenvironment{talign}
 {\csname align\endcsname}
 {\endalign}

\def\balignst#1\ealignst{\begin{talign*}#1\end{talign*}}
\def\balignt#1\ealignt{\begin{talign}#1\end{talign}}

\let\originalleft\left
\let\originalright\right
\renewcommand{\left}{\mathopen{}\mathclose\bgroup\originalleft}
\renewcommand{\right}{\aftergroup\egroup\originalright}

\def\tinycitep*#1{{\tiny\citep*{#1}}}
\def\tinycitealt*#1{{\tiny\citealt*{#1}}}
\def\tinycite*#1{{\tiny\cite*{#1}}}
\def\smallcitep*#1{{\scriptsize\citep*{#1}}}
\def\smallcitealt*#1{{\scriptsize\citealt*{#1}}}
\def\smallcite*#1{{\scriptsize\cite*{#1}}}

\def\mbf#1{\mathbf{#1}}
\def\mbb#1{\mathbb{#1}}

\def\mrm#1{\mathrm{#1}}

\def\R{\mathbb{R}}

\def\<{\left\langle} 
\def\>{\right\rangle}

\def\P{\mbb{P}} 

\ifdefined\nonewproofenvironments\else
\ifdefined\ispres\else

\newenvironment{proof-sketch}{\noindent\textbf{Proof Sketch}
  \hspace*{1em}}{\qed\bigskip\\}
\newenvironment{proof-idea}{\noindent\textbf{Proof Idea}
  \hspace*{1em}}{\qed\bigskip\\}
\newenvironment{proof-of-lemma}[1][{}]{\noindent\textbf{Proof of Lemma {#1}}
  \hspace*{1em}}{\qed\\}
\newenvironment{proof-of-theorem}[1][{}]{\noindent\textbf{Proof of Theorem {#1}}
  \hspace*{1em}}{\qed\\}
\newenvironment{proof-attempt}{\noindent\textbf{Proof Attempt}
  \hspace*{1em}}{\qed\bigskip\\}

\fi

\crefname{appendix}{App.}{App.}
\crefname{equation}{}{}
\crefname{lemma}{Lem.}{Lems.}
\crefname{claim}{Claim}{Claims}
\crefname{theorem}{Thm.}{Thms.}
\crefname{Corollary}{Cor.}{Cors.}
\crefname{algorithm}{Alg.}{Algs.}
\crefname{example}{Ex.}{Exs.}
\crefname{section}{Sec.}{Secs.}
\crefname{table}{Tab.}{Tabs.}
\crefname{remark}{Rem.}{Rems.}
\crefname{customthm}{Thm.}{Thms.}
\crefname{definition}{Def.}{Defs.}
\crefname{Proposition}{Prop.}{Props.}
\crefname{myremark}{Rem.}{Rems.}
\crefname{mylemma}{Lem.}{Lems.}
\crefname{mydefinition}{Def.}{Defs.}
\crefname{myproposition}{Prop.}{Props.}
\crefname{mycorollary}{Cor.}{Cors.}
\crefname{assumption}{Assum.}{Assums.}
\crefname{figure}{Fig.}{Figs.}
\crefname{proof}{Pf.}{Pfs.}
\crefname{enumi}{}{}
\crefname{name}{}{} 

\newcommand{\mcar}{\texttt{MCAR}\xspace}
\newcommand{\mar}{\texttt{MAR}\xspace}
\newcommand{\colmar}{\texttt{Col}-\mar}
\newcommand{\mnar}{\texttt{MNAR}\xspace}
\newcommand{\mnarself}{\texttt{Self-MNAR}\xspace}
\newcommand{\polarmnar}{\texttt{Pol-MNAR}\xspace}
\newcommand{\softpolarmnar}{\texttt{Soft-Pol-MNAR}\xspace}
\newcommand{\latentmnar}{\texttt{Latent-MNAR}\xspace}
\newcommand{\clustermnar}{\texttt{Clust-MNAR}\xspace}
\newcommand{\twophasemnar}{\texttt{Phase-MNAR}\xspace}
\newcommand{\censormnar}{\texttt{Censor-MNAR}\xspace}
\newcommand{\nnmar}{\texttt{NN-MNAR}\xspace}
\newcommand{\marblockneural}{\texttt{Block-MNAR}\xspace}
\newcommand{\method}{TabImpute\xspace}
\newcommand{\benchmark}{MissBench\xspace}

\newcommand{\numothermethods}{14\xspace}
\newcommand{\numpatterns}{11\xspace}

\author{Jacob Feitelberg\\Columbia University \And Dwaipayan Saha\\Columbia University \And Kyuseong Choi\\Cornell University \And Zaid Ahmad\\Columbia University \And Anish Agarwal\\Columbia University \And Raaz Dwivedi\\Cornell University }

\title{
One Pipeline, Many Transformers: Pattern-Specific Imputation Specialists for Tabular Missing Data
}

\begin{document}

\maketitle

\begin{abstract}

Missing data in tabular datasets forces practitioners into a hard choice: deploy a general-purpose imputer that may perform poorly for the problem at hand, or wait for someone to design a specialized algorithm. This problem is worsened by the fact that real-world missingness rarely satisfies the textbook missing completely at random (MCAR) assumption, as entries are often missing not at random (MNAR), where the probability of being observed depends on the missing data itself. We collapse this trade-off into a single pre-training pipeline that builds transformer-based imputation specialists through three components: an entry-wise featurization that recasts imputation as supervised prediction over row--column context, a synthetic data generator with pluggable missingness modules, and prior-data fitting on millions of synthetic tables. Swapping only the missingness module, with no changes to architecture, loss, or training, yields a pattern-specific specialist that outperforms methods purpose-built for that pattern. We validate this on MissBench, a new benchmark of 42 OpenML datasets and \numpatterns missingness patterns (including 9 MNAR variants) spanning medicine, finance, and engineering. Remarkably, training exclusively on MCAR yields a default model---TabImpute---robust across all tested patterns. In addition, the pattern-specific specialists produced by our pipeline outperform the 14 established baselines on their target patterns. We open-source the pipeline, models, and benchmark.

\end{abstract}

\section{Introduction}\label{sec:intro}
Missing data is ubiquitous across tabular datasets, affecting statisticians, economists, health officials, and businesses. For example, healthcare datasets may have unrecorded blood pressure measurements, or datasets merged from multiple sources may only share some features. Regardless of the source, missing data must usually be imputed to numerical values before employing statistical or machine learning models. Imputation methods range from simple approaches (e.g., constant values and averages) to more sophisticated techniques such as nearest-neighbor methods \citep{Batista01052003}, matrix factorization methods such as SoftImpute \citep{hastie2015matrix}, and random forest regression, notably the MissForest algorithm \citep{missforest2011}. However, each method in the literature is typically tailored for specific missingness settings, with performance varying significantly across datasets, domains, and missingness patterns \citep{van2012flexible,jarrett2022hyperimpute,agarwal2023causal,Ibrahim01032005}. When practitioners encounter a new missingness pattern, they must either default to a general-purpose method that may underperform in their setting, or wait for researchers to design a specialized algorithm, a process that requires significant expertise and effort. 

Building on recent advances in tabular representation learning \citep{hollmann2023tabpfn, ye2025closer}, we propose a flexible pre-training pipeline that eliminates this trade-off: by combining an entry-wise featurization with a synthetic data generator whose missingness module is interchangeable, our pipeline produces zero-shot transformer-based imputers that are both accurate on specific missingness patterns and robust across diverse settings, without requiring any change to the model architecture or training algorithm. We call the models trained under our pipeline TabImpute-P where P is a specific missingness pattern, and we refer to the default \mcar-trained TabImpute model as either TabImpute (default) or TabImpute for short. 

\cite{rubin1976inference} proposed analyzing missingness based on its relationship with the data-generating process to determine whether missingness biases downstream estimation. Rubin demonstrated that when missingness is independent of the underlying data, the observed data distribution provides an unbiased basis for estimation. This framework categorizes missingness into three classes: Missing Completely At Random (\mcar), Missing At Random (\mar), and Missing Not At Random (\mnar) \citep{van2012flexible, sportisse2020aimputation, sportisse2020b}. \mcar defines scenarios where missingness is independent of all data values, both observed and unobserved. \mar encompasses cases where missingness depends on observed variables that can be appropriately conditioned upon during analysis. \mnar describes situations where missingness depends on unobserved factors that cannot be easily conditioned on.

In practice, distinguishing \mar from \mnar missingness using the observed data alone is generally impossible because the distinction depends on unobserved values that are, by definition, unavailable \citep{little2019statistical, allison2002}. This problem is especially acute in small datasets, where tests for the missingness patterns, such as Little's chi-squared test \citep{little1988test} for \mcar, have low statistical power. 

\begin{figure*}[!h]
    \centering
    \includegraphics[width=\linewidth]{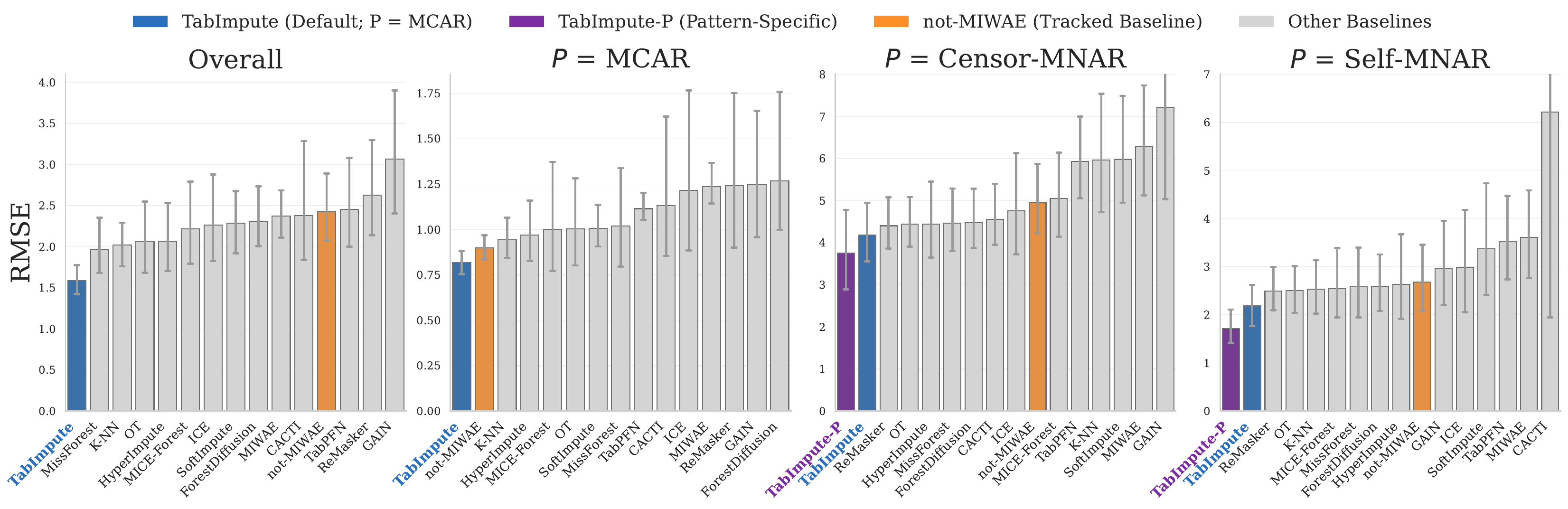}
    \caption{\textbf{Evaluation on real-world OpenML small datasets: MissBench.} We compare the \method models (blue and purple) with \numothermethods popular baseline methods on \benchmark. The leftmost panel shows root mean squared error (RMSE) averaged over all 11 patterns, followed by RMSE for \mcar, \censormnar, and \mnarself. We highlight not-MIWAE as an example of the high variance that the baseline methods have across missingness patterns. Specialized TabImpute models are labeled TabImpute-P where P is the specific pattern they are trained for. The default TabImpute model is trained on \mcar, so only one TabImpute model is shown in that panel. RMSE is calculated for each method, standardized within a dataset, and averaged across datasets. Error bars indicate 95\% confidence intervals, computed across datasets, not across missing entries or dataset-pattern pairs. See \cref{fig:all-bar-plot} in the Appendix for the full set of patterns.}
    \label{fig:results}
\end{figure*}

This creates two distinct practical scenarios. In the first, the practitioner has no knowledge of the missingness mechanism and requires an imputer that is accurate and robust across a wide range of possible patterns. In the second, the practitioner has domain knowledge suggesting a specific mechanism and would like the imputation method to exploit this knowledge for improved accuracy. Existing methods often address only one of these scenarios: general-purpose methods like MICE and MissForest make no use of known mechanism structure, while pattern-specific methods like not-MIWAE \citep{ipsen2020not} require the practitioner to correctly specify a missingness model and do not transfer well to other patterns. Our pipeline addresses both scenarios within a single framework. When the mechanism is unknown, the pipeline's default model provides robust performance across diverse patterns (see \cref{fig:results}). When the mechanism is known, the practitioner simply specifies it in the synthetic data generator to produce a specialized model, without designing a new algorithm.

\begin{figure*}
    \centering
    \includegraphics[width=\textwidth]{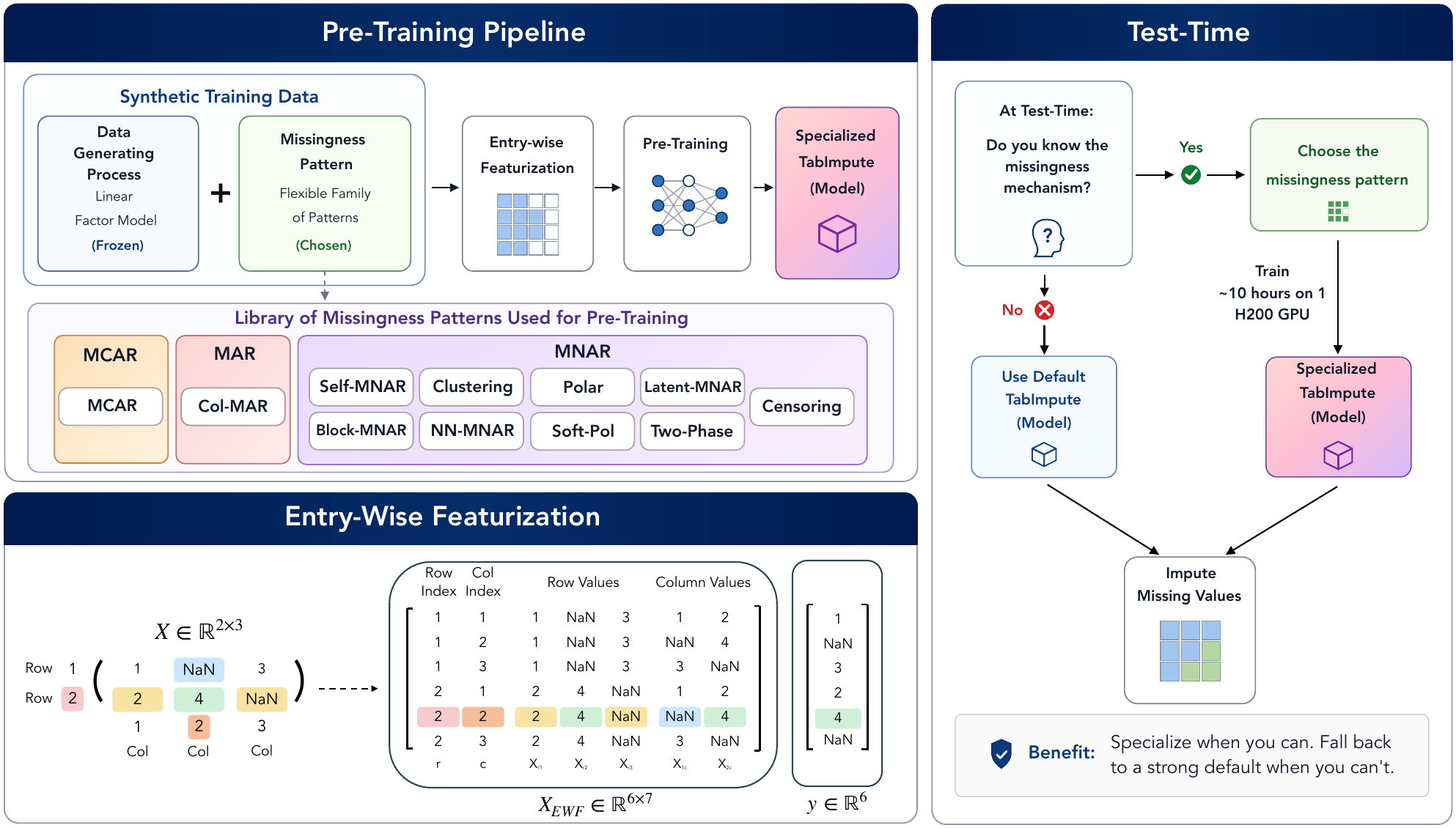}
    \caption{\textbf{Our pipeline.} (Left) The pre-training pipeline combines a linear factor model with an interchangeable missingness module to generate synthetic training data, which is processed through entry-wise featurization and pre-trained to produce a TabImpute model. Only the missingness module changes; the architecture and training algorithm remain fixed. (Right) At test time, if the missingness mechanism is known, the practitioner uses the corresponding specialized model or trains their own TabImpute model (around 10 hours on an H200 GPU); otherwise, the default MCAR-trained model provides robust performance across diverse patterns. (Bottom left) Entry-wise featurization transforms each entry of a table into a row of features comprising its row index, column index, row context, and column context, enabling parallel prediction of all missing entries. The highlighted row illustrates how entry (2,2) is featurized.}
    \label{fig:overview}
\end{figure*}

The main contributions of this work can be summarized as follows (also shown in \cref{fig:overview}):
\begin{enumerate}
    \item We propose a flexible pre-training pipeline for tabular missing data imputation that addresses both practical scenarios: when the missingness mechanism is unknown, the pipeline produces a robust default imputer, and when the mechanism is known, the practitioner specifies it in the synthetic data generator to produce a specialized imputer, all without modifying the model architecture or training algorithm. The pipeline combines a new entry-wise featurization (EWF), which recasts imputation as a supervised prediction problem with a 100$\times$ speedup over prior column-wise TabPFN-based imputation (see \cref{fig:runtime} in \cref{sec:empirical}), with a synthetic data generator whose missingness module is interchangeable (see \cref{sec:model-arch} and \cref{sec:syn-data} for details).
    \item We demonstrate that specialized models produced by the pipeline outperform existing methods designed for specific patterns, while the default model, trained on \mcar alone, is robust across \numpatterns diverse patterns including 9 structured \mnar patterns. This eliminates the need for practitioners to either design new algorithms for their missingness setting or settle for a general-purpose method that may underperform (see \cref{sec:syn-data}, \cref{app:mnar}, and \cref{tab:comprehensive_missingness_hyperparams} for details).
    \item We introduce \benchmark, a comprehensive benchmark of 42 small, real-world OpenML \citep{vanschoren2014openml} datasets and \numpatterns missingness patterns spanning biology, medicine, finance, and education, and demonstrate the pipeline's strong performance against \numothermethods established baselines across both scenarios (see \cref{sec:empirical} for details).
\end{enumerate}
Our code, trained TabImpute models, and implementation details are available on GitHub.\footnote{\url{https://github.com/jacobf18/tabular}}

\subsection{Previous Work}

We primarily build on missing-data imputation and tabular representation learning (TRL) literature. While missing-data imputation is well-studied, with established theory dating back to the 1970s \citep{rubin1976inference}, TRL using tabular foundation models is relatively new \citep{muller2022transformers,zhang2025tabpfnmodelruleall,hollmann2023tabpfn,hollmann2025accurate}. Below, we describe relevant prior work.

\paragraph{Imputation methods.} Given the widespread nature of missing data, numerous imputation techniques have been proposed. These include linear models over columns \citep{ice2011}, random forest models \citep{missforest_python}, ensemble methods \citep{jarrett2022hyperimpute}, nearest neighbor-based methods \citep{chin2025n}, and generative adversarial networks \mbox{\citep{gain2018}}. HyperImpute \citep{jarrett2022hyperimpute} combines the power of multiple classical imputation methods through iterative imputation (IM). IM loops over each column, using other columns to predict missing values until convergence. 

\paragraph{Tabular representation learning.} Tabular representation learning focuses on building models that can generalize across diverse tabular domains. The pioneering work in this area is TabPFN, proposed in \citep{hollmann2023tabpfn} and improved in \citep{hollmann2025accurate}. 
Since TabPFN's introduction, numerous variants have emerged to address scalability and performance limitations, as well as further work aimed at clarifying its internal representations \citep{zhang2025tabpfnmodelruleall,ye2025closer}: TabICLv2 \citep{qu2025tabicl} is a highly scalable tabular foundation model, MITRA \citep{MITRA} is a model pre-trained on purely synthetic data from a mix of random classifiers/regressors, and TabDPT \citep{ma2024tabdpt} is a tabular model pre-trained on real-world data. 
These models demonstrate that pre-training can improve performance on real-world tabular data, motivating our synthetic pre-training pipeline for imputation.

\section{Background on Prior-data Fitted Networks and TabPFN}

Prior-data Fitted Networks (PFNs) are a class of models that learn to approximate Bayesian inference for a given prior \citep{muller2022transformers}. Instead of fitting a new model from scratch, a PFN is a single large pre-trained Transformer that performs classification or regression in a single forward pass. This process, known as in-context learning (ICL), allows the model to make predictions using sequences of labeled examples provided directly in the input, without requiring any gradient updates \citep{dong2024surveyincontextlearning}. The entire prediction algorithm is contained in the weights of the network, which is trained once on millions of synthetically generated datasets sampled from the prior. At inference time, the trained PFN takes a real-world dataset, composed of context and query samples, as a set-valued input and returns a distribution over the output space. This output space is categorical for classification tasks and the real line for regression tasks.

PFNs are rooted in Bayesian supervised learning, where the primary objective is to model the posterior predictive distribution (PPD) \citep{mackay1992practical,seeger2004gaussian,blei2017variational}. Since computing the PPD is often intractable \citep{mackay1992practical}, PFNs instead learn to approximate the PPD offline through a process called synthetic prior fitting \citep{muller2022transformers}. This is achieved using a prior specified by a sampling scheme that first samples a data-generating mechanism, $\phi \sim p(\phi)$, and then samples a synthetic dataset, $D \sim p(D|\phi)$. This process is repeated to generate millions of diverse datasets for training. The network's parameters, $\theta$, are then optimized to predict held-out test samples ($D_{\text{test}} \subset D$) conditioned on the rest of the dataset ($D_{\text{train}} = D \setminus D_{\text{test}}$). The training objective is to minimize the negative log likelihood (NLL) loss on these held-out examples:
$
    \mathcal{L}(\theta) = \mathbb{E}_{((x_{\text{test}}, y_{\text{test}}) \cup D_{\text{train}}) \sim p(D)}[-\log q_{\theta}(y_{\text{test}} | x_{\text{test}}, D_{\text{train}})].
$
Minimizing this loss ensures that the trained neural network $q_{\theta}$ approximates the true Bayesian PPD for the specified prior.

TabPFN is a PFN built specifically for tabular supervised learning \citep{hollmann2023tabpfn}. The model employs a novel two-way attention mechanism specifically designed for tabular data. 
The architecture uses alternating attention patterns: each cell first attends to other features within its row (inter-feature attention), then attends to the same feature across all rows (inter-sample attention). This design ensures permutation invariance for both samples and features while enabling efficient scaling to larger tables than those seen during training. TabPFN v2 \citep{hollmann2025accurate} retains the core training paradigm of TabPFN v1 while introducing several key enhancements that improve accuracy, runtime, and applicability. Going forward, when referring to TabPFN, we specifically mean TabPFN-2.6, the most recent version of TabPFN. The team behind TabPFN also created an imputation method in their \texttt{tabpfn-extensions} Python package by applying TabPFN iteratively in a column-wise imputation scheme, which we refer to simply as TabPFN in this paper (available on GitHub\footnote{\url{https://github.com/PriorLabs/tabpfn-extensions/blob/main/src/tabpfn_extensions/unsupervised/unsupervised.py}}). 

\section{Pre-Training Imputers on Synthetic Data}\label{sec:arch}

We pre-train the models on our novel missing-data priors using the entry-wise featurization described below. Training and evaluation used an H200 GPU and an Intel Xeon Platinum 8592+ CPU. Training each model took approximately 10-15 hours, depending on the runtime of the missingness generator component.


\paragraph{Entry-wise featurization.}\label{sec:model-arch}Let $X^*$ be the complete matrix with $m$ rows and $n$ columns, $\Omega$ be the set of missing entry indices, and $X$ be the matrix with induced missingness: for each entry, $X_{ij} = X^*_{ij}$ if entry $(i,j) \notin \Omega$ (observed) and $X_{ij}=\star$ if $(i,j) \in \Omega$ (missing) where $\star$ denotes a missing entry. Let $\Omega_{\text{obs}} = [m] \times [n] \setminus \Omega$. Each row of the entry-wise feature matrix is $(i \oplus j \oplus X_{i, :} \oplus X_{:, j})$ for entry  $i,j \in [m] \times [n]$, where $X_{i, :}$ denotes the $i$-th row, $X_{:, j}$ the $j$-th column, and $\oplus$ concatenation. Each row's target is $y_{ij}=X_{ij}$. During pre-training, we train the model to predict target values for all $(i,j) \in \Omega$. This procedure creates a feature matrix with $mn$ rows and $m+n+2$ features, including row and column indices. This featurization captures all necessary information for each cell through its row and column context while enabling parallel computation of missing entries on GPUs. Although the input matrix size increases, the gains from parallelization outweigh this cost for small tables. Note that the features are created using the matrix with missing values $X$, not the true matrix $X^*$. 

We demonstrate entry-wise featurization on a small toy matrix with missing values in the bottom-left panel in \cref{fig:overview}. During pre-training, we have access to the true values of the $\mathrm{NaN}$s in $y$. However, these values are only used to update the model's weights via the loss function; the true values are never input into the model. The model internally handles $\mathrm{NaN}$ inputs by replacing their values with the column mean in the input matrix and then appending a binary matrix indicating which values are $\mathrm{NaN}$. Internally, the $\mathrm{NaN}$ values in $y$ are replaced with the mean of the non-$\mathrm{NaN}$ values (observed values) in $y$.

\begin{figure}[t]
\centering

\includegraphics[width=\textwidth]{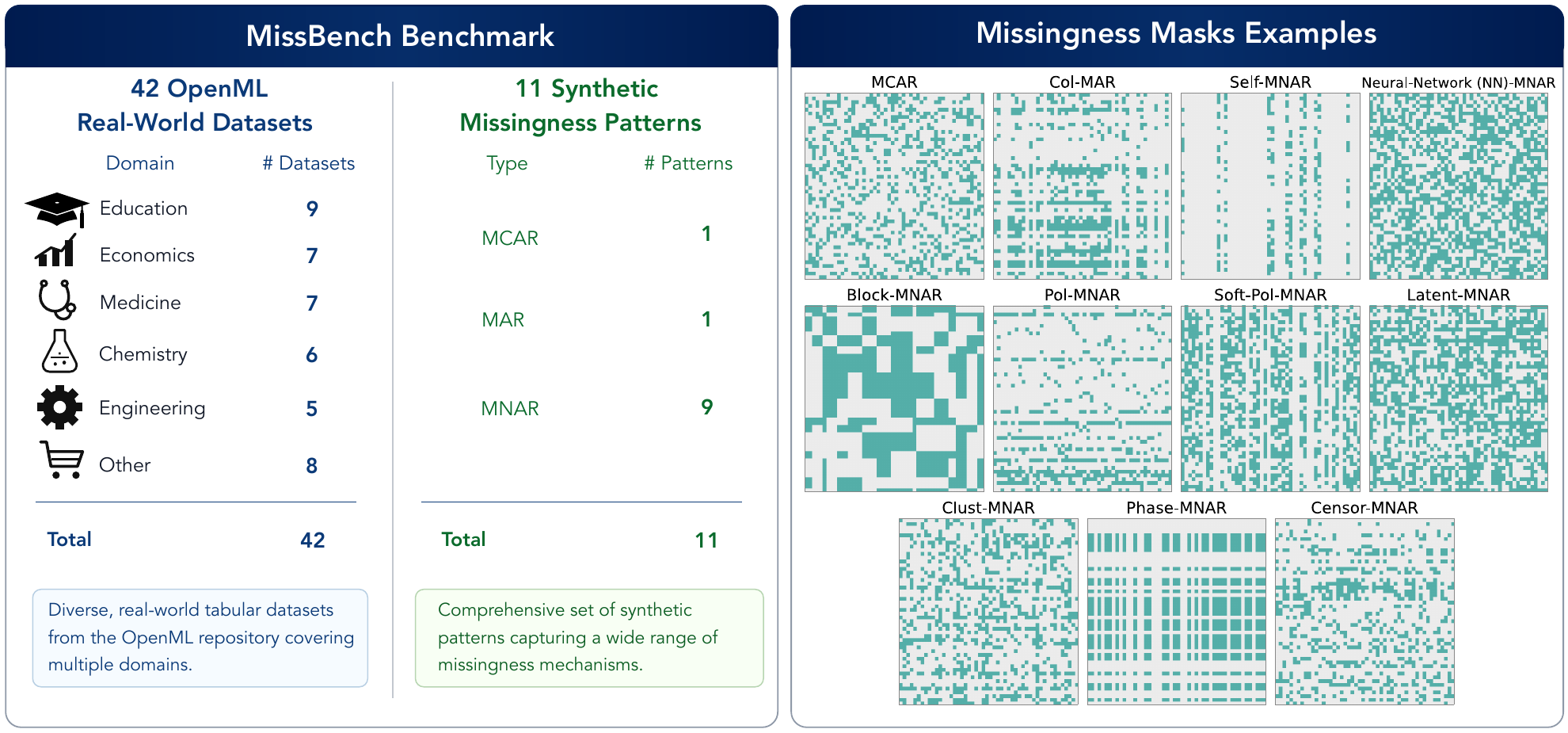}

\caption{\textbf{Missngness Patterns and MissBench.} (Left) We introduce MissBench, a benchmark of 42 real-world OpenML datasets spanning multiple domains with 11 missingness patterns including 9 \mnar patterns. (Right) Examples of missingness masks from MissBench. Blue entries are unobserved; gray entries are observed. The pipeline's interchangeable missingness module generates these and other patterns during synthetic pre-training.}
\label{fig:toy-ewf-and-missingness}
\end{figure}

\paragraph{Architecture.} Following TabPFN's terminology, we refer to rows in the feature matrix with observed target values (non-$\mathrm{NaN}$) as \emph{input} points and rows with missing values as \emph{query} points. We utilize an architecture similar to TabPFN v2's transformer architecture with one major change: we remove the attention mask that restricts information flow between query points. In the original supervised learning setting considered by TabPFN, this mask encodes an inductive bias that predictions for query points should be conditionally independent, reflecting the standard assumption that test points do not influence each other’s predictions. In contrast, in missing-data imputation, there is no assumption that missing entries (query points) should be independent. Indeed, under many missingness mechanisms, particularly non-\mcar settings, missing values may be strongly correlated through shared latent structure. Allowing query points to exchange information, therefore, better reflects the structure of the imputation problem.

\subsection{Synthetic Training Data Generation}\label{sec:syn-data}
For each model's training, we generate around $1.2$ million synthetic tables with missing values through a two-step process: first, generating underlying data, then introducing missingness patterns on top. We detail both procedures below.

\subsubsection{Data Generation: Linear Factor Models}

We generate synthetic data using a linear factor model (LFM) \citep{linearfactor}. LFMs are commonly used in matrix completion literature to prove error bounds for matrix completion algorithms \citep{netflixprize,rechtexact}. This family of models generates a data matrix $Y \in \mathbb{R}^{m \times n}$ by assuming the data lies on or near a low-dimensional subspace. The simplest case generates the data matrix $Y$ as the inner product of two lower-rank latent factor matrices, $U \in \mathbb{R}^{m \times k}$ and $V \in \mathbb{R}^{n \times k}$, where $k \ll \min(n, m)$ is the rank: $Y = UV^T$. The latent vectors (rows of $U$ and $V$) are sampled from a variety of distributions including Gaussian, Laplace, and Student's-t.

During pre-training, we also experimented with using nonlinear factor models. The model trained on LFMs performed the best, as shown in \cref{tab:errors_with_nonlinear} in \cref{app:add-tbl}. However, the TabImpute model trained with a nonlinear factor model attained almost the same RMSE as the LFM TabImpute model, so we expect that future work on the data-generating part of the pipeline could further improve the generalizability of the models.

\subsubsection{Missingness Patterns}

After generating a complete data matrix $X^* \in \mathbb{R}^{m \times n}$, we introduce missingness by applying a mask $M \in \{0, 1\}^{m \times n}$, where the entry value $M_{ij}=1$ if $(i, j) \in \Omega_{\mrm{obs}}$. To ensure \method is robust and generalizable to the variety of ways data can be missing in real-world scenarios, we pre-train on millions of synthetic datasets, each with a different mask. For convenience, we define $p_{ij} = \P(M_{ij} = 1)$, the propensity of each entry. We evaluate TabImpute on \numpatterns different missingness patterns:  \mcar, $1$ \mar pattern, and $9$ \mnar patterns. For examples of these, see the right panel in \cref{fig:toy-ewf-and-missingness}.

\paragraph{\mcar:} \mcar missingness means the probability of an entry being missing is independent of all data values, both observed and unobserved. In our implementation, we draw the missingness indicators $M_{ij}$ i.i.d. from a Bernoulli distribution $M_{ij} \sim \mrm{Bern}(p)
$ across all $(i,j)\in [m]\times [n]$ for some constant $p \in (0, 1)$. This is the simplest form of missingness, but is unrealistic \citep{van2012flexible}.

\paragraph{\mar:} For \mar missingness, the probability of an entry being missing depends only on the observed values $X$. In other words, the randomness in \mar can be explained by conditioning on observed factors. Additionally, every entry has a positive probability of being observed~(i.e., $p_{ij} > 0$). We simulate \mar through column-wise \mar, denoted \colmar: we choose several columns as predictor columns and use those values to mask entries in other columns. This is similar to the \mar approach taken in \citep{jarrett2022hyperimpute}.

\paragraph{\mnar:} For \mnar, the most complex missingness class, the probability of an entry being missing can depend on unobserved factors. Note that \mnar patterns are significantly more difficult to handle systematically, often requiring specialized methods for a specific kind of \mnar pattern \citep{van2012flexible}. Due to the flexibility of our entry-wise featurization, \method can produce imputations for these highly complex scenarios. HyperImpute was tested on two \mnar patterns in the Appendix of \citep{jarrett2022hyperimpute}, one where values are further masked after an \mar pattern and another where values outside a certain range are masked. We build on this work by testing on $9$ \mnar patterns. For example, we utilize the expressiveness of neural networks to create \mnar patterns with random propensity functions (\nnmar and \marblockneural) and censoring where sensor readings fall outside a detectable range (\censormnar). For details on each method, see \cref{app:mnar} and \cref{tab:comprehensive_missingness_hyperparams}.

\subsection{Training the TabImpute models}\label{sec:training}

We use the negative log likelihood (NLL) loss and Riemann distribution output, both proposed in \citep{muller2022transformers}. Since we can generate an unlimited amount of synthetic data, each batch of synthetic data passed to the model during training is new. This allows our model to learn the underlying data-generating process and missingness mechanisms without risk of memorization. For each TabImpute model, we use the AdamW optimizer \citep{loshchilov2017fixing}, a learning rate of 3e-4, a weight decay of 1e-2, a batch size of $16$, and around $1.2$ million synthetic datasets.

We trained the default TabImpute model only on \mcar missingness, finding that this model was the most robust across all patterns. We had initially attempted to train the model sequentially on one missingness pattern at a time, but found that the network always experienced \emph{catastrophic forgetting} \citep{mccloskey1989catastrophic} irrespective of learning rate (i.e., it forgot how to handle the previous missingness patterns). After this, we attempted to mix several missingness patterns together: \mcar, \colmar, and \mnarself. However, the model's performance degraded overall because it focused only on \mnarself. We attempted several techniques, including weighting schedules and GradNorm \citep{chen2018gradnorm}, a gradient normalization technique for multi-task learning, but found them unsuccessful in improving performance.

\paragraph{Why \mcar pre-training produces a robust default model.} By generating millions of synthetic datasets under \mcar, the default model is exposed to a broad and diverse set of missingness masks. While \mar and \mnar mechanisms induce biased mask distributions that emphasize specific structural patterns, many such masks overlap with the support of masks encountered under \mcar generation, but with different frequencies. Importantly, models produced by our pipeline do \emph{not} model the underlying missingness dynamics, but rather the conditional data distribution given the observed data, $P(X_{ij}|X_{\Omega_{\text{obs}}})$. This makes the \mcar-trained default model well suited as a general-purpose imputer that achieves strong predictive accuracy across many missingness patterns without pattern-specific tuning. However, when the missingness mechanism is known a priori, the pipeline's flexibility becomes its key advantage: by simply changing the missingness module in the synthetic data generator, without modifying the model architecture or training algorithm, practitioners can train specialized models that outperform the default, as shown in \cref{tab:specialized_models_comparison} and \cref{fig:results}. This eliminates the need to design new imputation algorithms for each missingness pattern.

\begin{table*}[!ht]
\centering
\caption{\centering RMSE $\pm$ 95\% confidence interval by missingness pattern. \\Confidence intervals are computed across the 42 datasets, not across missing entries or dataset-pattern pairs. For the full table, see \cref{tab:normalized_negative_rmse_by_pattern_full}.}\label{tab:normalized_negative_rmse_by_pattern}
\resizebox{0.9\textwidth}{!}{\begin{tabular}{llccccc}
\toprule
& Pattern & TabImpute & MissForest & K-NN & OT & HyperImpute \\
\midrule
\textbf{\mcar} & \mcar & \textbf{0.819 ± 0.065} & 1.018 ± 0.291 & 0.943 ± 0.119 & 1.004 ± 0.279 & 0.969 ± 0.184 \\
\noalign{\vskip 1.5pt}
\hdashline
\noalign{\vskip 1.5pt}
\textbf{\mar} & \colmar & 1.076 ± 0.164 & 1.197 ± 0.229 & \textbf{1.075 ± 0.082} & 1.377 ± 0.340 & 1.117 ± 0.228 \\
\noalign{\vskip 1.5pt}
\hdashline
\noalign{\vskip 1.5pt}
\multirow{9}{*}{\textbf{\mnar}} & \mnarself & \textbf{2.184 ± 0.435} & 2.580 ± 0.716 & 2.539 ± 0.555 & 2.500 ± 0.499 & 2.636 ± 0.949 \\
& \nnmar & \textbf{0.854 ± 0.083} & 1.070 ± 0.300 & 1.004 ± 0.194 & 1.065 ± 0.310 & 1.114 ± 0.310 \\
& \marblockneural & \textbf{0.939 ± 0.051} & 1.430 ± 0.752 & 1.039 ± 0.161 & 1.489 ± 0.721 & 1.325 ± 0.656 \\
& \polarmnar & 2.680 ± 1.318 & 4.433 ± 3.042 & 3.660 ± 1.755 & 5.290 ± 4.726 & 5.492 ± 4.183 \\
& \softpolarmnar & \textbf{2.051 ± 0.515} & 2.498 ± 0.828 & 3.157 ± 1.428 & 2.584 ± 1.038 & 2.713 ± 0.901 \\
& \latentmnar & \textbf{0.867 ± 0.060} & 1.052 ± 0.276 & 0.958 ± 0.070 & 0.975 ± 0.169 & 1.082 ± 0.257 \\
& \clustermnar & \textbf{0.872 ± 0.075} & 0.947 ± 0.111 & 0.988 ± 0.120 & 0.965 ± 0.131 & 0.917 ± 0.082 \\
& \twophasemnar & 0.917 ± 0.058 & 0.937 ± 0.061 & \textbf{0.915 ± 0.055} & 1.022 ± 0.067 & 0.931 ± 0.095 \\
& \censormnar & \textbf{4.181 ± 0.721} & 4.465 ± 0.759 & 5.961 ± 1.531 & 4.442 ± 0.622 & 4.444 ± 0.981 \\
\midrule
& Overall & \textbf{1.585 ± 0.174} & 1.966 ± 0.325 & 2.022 ± 0.285 & 2.065 ± 0.456 & 2.067 ± 0.423 \\
\bottomrule
\end{tabular}}
\end{table*}

\section{Results on OpenML Datasets: \benchmark}\label{sec:empirical}

To evaluate \method against other methods, we introduce \benchmark : a missing-data imputation benchmark using $42$ OpenML \citep{OpenML2013} small tabular datasets with \numpatterns synthetic missingness patterns (see the right panel of \cref{fig:toy-ewf-and-missingness}). For every dataset and missingness pattern, we test each method's ability to impute masked values. The $42$ OpenML datasets span domains such as medicine, engineering, and education. We evaluate on datasets with numerical values without pre-existing missingness before applying synthetic patterns. Each dataset with its description is listed in \cref{tab:openml_datasets} in \cref{app:add-tbl}. The missingness patterns include \mcar, \colmar, and $9$ different \mnar patterns, with details provided in \cref{app:mnar}. We test a suite of imputation methods on \benchmark: SoftImpute \citep{hastie2015matrix}, MissForest \citep{missforest2011}, iterative chained estimators (ICE/MICE) \citep{ice2011,royston2011multiple}, GAIN \citep{gain2018}, MIWAE \citep{mattei2019miwae}, not-MIWAE \citep{ipsen2020not}, an optimal transport-based method \citep{muzellec2020missing}, ForestDiffusion \citep{jolicoeur2024generating}, ReMasker \citep{du2023remasker}, CACTI \citep{gorla2025cacti}, and TabPFN (see \cref{sec:intro}). 

\begin{figure}
    \centering
    \begin{tabular}{cc}
         \includegraphics[width=0.4\textwidth]{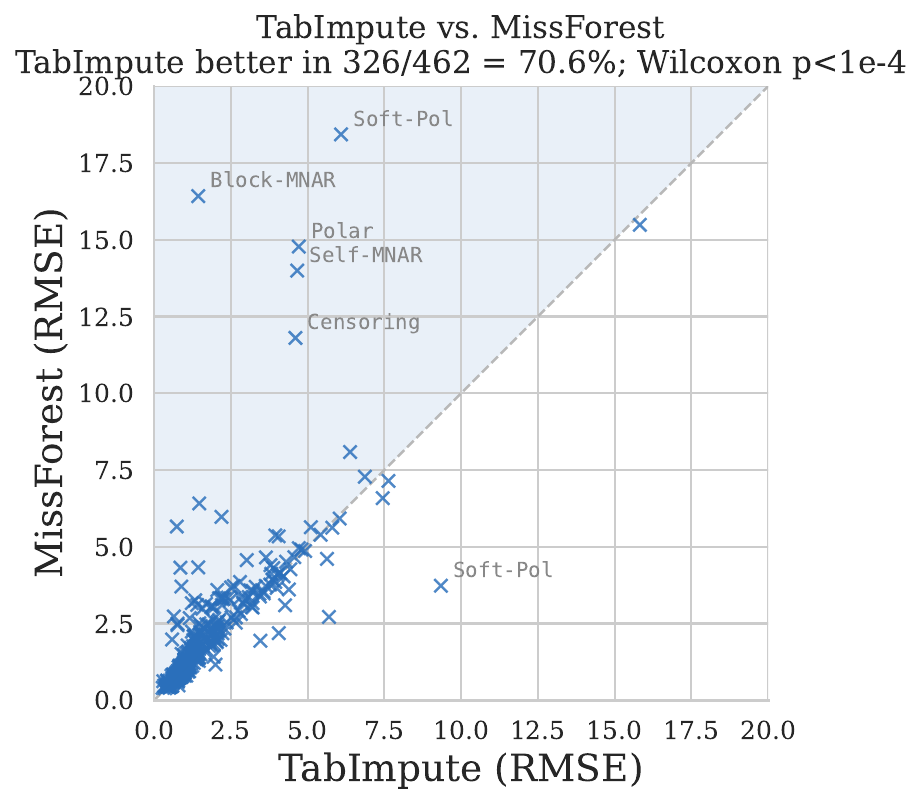} & \includegraphics[width=0.36\textwidth]{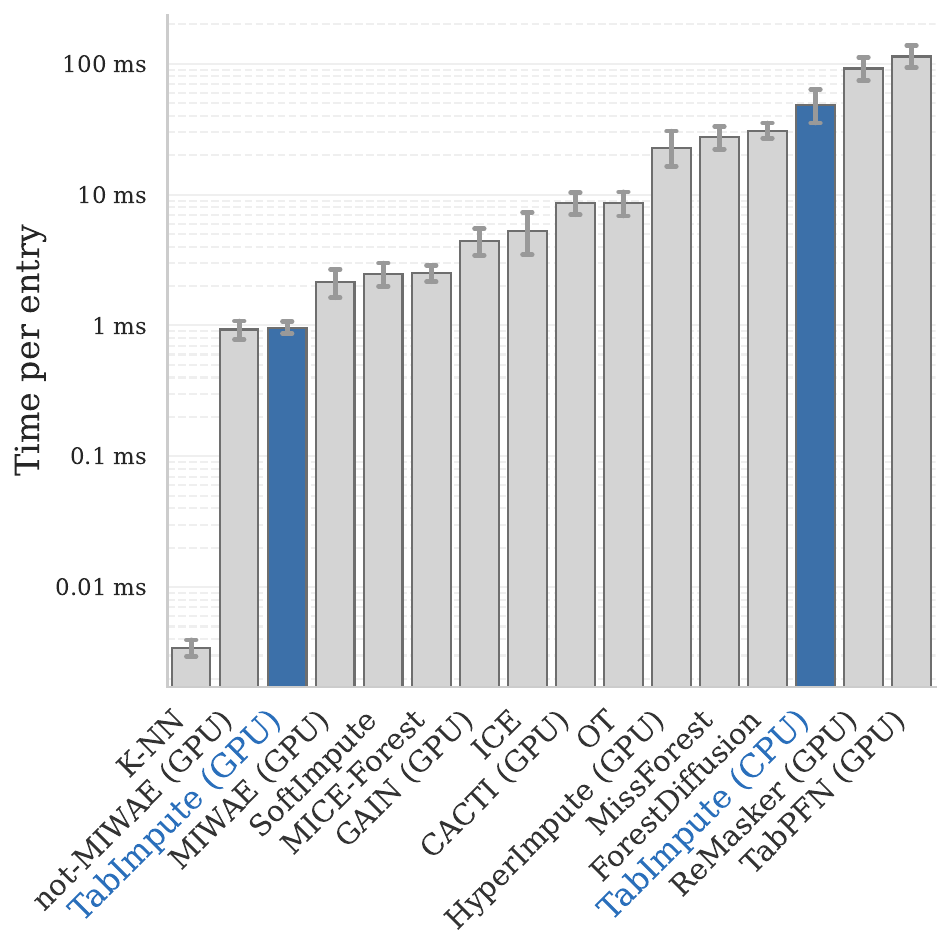}\\
        (a) RMSE comparison with MissForest & (b) Runtime
    \end{tabular}
    \caption{\textbf{TabImpute is both accurate and fast.} (a) Head-to-head comparison of the default TabImpute model against MissForest, the strongest baseline from prior literature. Each point represents one dataset–missingness pattern pair; points above the diagonal indicate TabImpute achieves lower RMSE. Points furthest from the diagonal are labeled with their corresponding missingness pattern. TabImpute wins on 70.6\% of comparisons (326/462, Wilcoxon $p <$1e-4). (b) Runtime per entry for all methods. Methods labeled (GPU) are run on an H200 GPU; all others run on CPU. The default TabImpute model achieves over 100× faster runtime than column-wise TabPFN.}
    \label{fig:runtime}
\end{figure}

Following \cite{missforest_python}, we report root mean squared error (RMSE) for each method and missingness pattern. RMSE is normalized by the standard deviation of the true values in each column: $\frac{1}{\sigma_{:,j}}{\big(\frac{1}{|\Omega|}\sum_{(i,j)\in \Omega} \big(X^{\text{true}}_{ij} - X^{\text{imputed}}_{ij}\big)^2\big)^{1/2}}$ where $\sigma_{:,j}$ is the standard deviation of the true values in column $j$. This is the same metric used in \cite{missforest2011}. \cref{tab:normalized_negative_rmse_by_pattern} presents results for each missingness pattern as well as overall performance. \method achieves the lowest overall RMSE and the lowest RMSE on 8 out of the 11 individual patterns (see the full results in \cref{tab:normalized_negative_rmse_by_pattern_full} in the Appendix). TabImpute also achieves the best column-wise $R^2$ (shown in \cref{tab:r_squared_by_pattern} in the Appendix), demonstrating that TabImpute also recovers structure at the whole-feature level. Finally, TabImpute's relative performance gets better as missingness increases (shown in \cref{fig:mcar-prob} in \cref{app:add-test}), which is expected since it leverages data from pre-training rather than relying solely on available dataset information.

\paragraph{Specialized \method Models.} 
\cref{tab:specialized_models_comparison} presents the performance of specialized TabImpute models, each trained on a different missingness pattern using our pipeline. The key finding is that, for the majority of patterns, the model trained on that specific pattern achieves top-2 performance without any modification to the model architecture or training algorithm. For example, the \mnarself, \censormnar, and \polarmnar models achieve the lowest RMSE on their corresponding patterns. Notably, the \mnarself-trained model also achieves top-2 performance on \censormnar, suggesting shared structure between these value-dependent missingness mechanisms. Two patterns, \nnmar and \latentmnar, are exceptions where the pattern-matched model does not rank in the top 2 among the specialized TabImpute models. We attribute this to our use of a relatively uniform optimization setup across all specialized models, with only minor learning rate adjustments; these two patterns likely require more careful hyperparameter tuning due to the complexity of their missingness generators. We expect that dedicated tuning would close this gap, and we leave this exploration to future work. However, as shown in the full set of plots in \cref{fig:all-bar-plot} in the Appendix, every specialized TabImpute model, including the ones for \nnmar and \latentmnar, is top-2 when only comparing with the default TabImpute and the \numothermethods baseline methods. Across all patterns, the \mcar-trained default model remains a strong baseline, achieving top-2 overall performance (1.585) and ranking first or second on 9 out of the \numpatterns individual patterns, confirming its value as a robust default when the missingness mechanism is unknown. Finally, not-MIWAE allows specifying a propensity model to make it specialized as well, which we do for \mnarself in \cref{fig:not-miwae} in the Appendix. Both our default TabImpute model and TabImpute-\mnarself outperform this specialized not-MIWAE.

\begin{table}[t]
\centering
\caption{\centering RMSE for specialized TabImpute models.\\
The top-2 models are bolded for each missingness pattern. See \cref{tab:specialized_models_comparison} in the Appendix for the full version of this table with 95\% confidence intervals. Note that the header \mnar patterns have the word \mnar removed for space constraints.
}
\label{tab:specialized_square}

\resizebox{\textwidth}{!}{\begin{tabular}{llc:c:ccccccccc}
\toprule
& & \textbf{\mcar} & \textbf{\mar} & \multicolumn{9}{:c}{\textbf{\mnar}}\\
\midrule
& Trained on $\Rightarrow$
& \shortstack{\mcar}
& \shortstack{\colmar}
& \shortstack{\texttt{Self}}
& \shortstack{\texttt{NN}}
& \shortstack{\texttt{Block}}
& \shortstack{\texttt{Pol}}
& \shortstack{\texttt{Soft-Pol}}
& \shortstack{\texttt{Latent}}
& \shortstack{\texttt{Clust}}
& \shortstack{\texttt{Phase}}
& \shortstack{\texttt{Censor}} \\
\midrule
\textbf{\mcar} & \mcar & \textbf{0.819} & \textbf{0.871} & 1.135 & 1.153 & 0.878 & 1.037 & 1.653 & 0.890 & 0.874 & 1.043 & 2.259 \\
\noalign{\vskip 1.5pt}
\hdashline
\noalign{\vskip 1.5pt}
\textbf{\mar} &\colmar & \textbf{1.076} & \textbf{1.072} & 1.770 & 1.442 & 1.214 & 1.358 & 2.471 & 1.163 & 1.082 & 1.125 & 2.526 \\
\noalign{\vskip 1.5pt}
\hdashline
\noalign{\vskip 1.5pt}
\multirow{9}{*}{\textbf{\texttt{\mnar}}}& \texttt{\mnarself} & \textbf{2.184} & 2.214 & \textbf{1.716} & 2.957 & 2.341 & 2.839 & 5.117 & 2.446 & 2.330 & 2.341 & 3.593 \\
& \texttt{\nnmar} & \textbf{0.854} & 0.925 & 1.070 & 0.888 & 0.934 & 1.039 & 1.808 & 0.908 & \textbf{0.886} & 1.042 & 2.014 \\
& \texttt{\marblockneural} & \textbf{0.939} & 1.017 & 1.304 & 1.303 & \textbf{0.985} & 1.071 & 2.268 & 1.007 & 0.994 & 1.056 & 2.288 \\
& \polarmnar & 2.680 & \textbf{1.767} & 2.780 & 2.961 & 2.065 & \textbf{1.316} & 7.547 & 2.096 & 2.456 & 3.479 & 8.324 \\
& \softpolarmnar & \textbf{2.051} & 2.159 & 2.391 & 2.458 & 2.095 & 2.207 & \textbf{1.953} & 2.207 & 2.057 & 2.717 & 2.465 \\
& \latentmnar & \textbf{0.867} & 0.940 & 1.097 & 1.097 & 0.928 & 1.043 & 1.566 & 0.915 & \textbf{0.894} & 1.038 & 2.083 \\
& \clustermnar & \textbf{0.872} & 0.922 & 1.084 & 1.086 & 0.903 & 1.032 & 1.377 & 0.895 & \textbf{0.887} & 1.027 & 1.921 \\
& \twophasemnar & \textbf{0.917} & 0.938 & 1.282 & 1.137 & 0.974 & 1.158 & 1.352 & 0.996 & 0.934 & \textbf{0.911} & 2.198 \\
& \censormnar & 4.181 & 4.220 & \textbf{3.018} & 5.425 & 4.427 & 5.066 & 7.971 & 4.638 & 4.483 & 5.247 & \textbf{3.758} \\
\midrule
& Overall & \textbf{1.585} & \textbf{1.549} & 1.695 & 1.991 & 1.613 & 1.742 & 3.189 & 1.651 & 1.625 & 1.911 & 3.039 \\
\bottomrule
\end{tabular}}
\end{table}

\paragraph{Comparison with other transformer-based methods.} For the other transformer-based methods, ReMasker \citep{du2023remasker} and CACTI \citep{gorla2025cacti}, we observe lower relative performance than reported in their original papers. This difference is likely attributable to a regime mismatch: both methods are designed to operate on much larger tables. For example, Fig. 9 in \citep{du2023remasker} shows ReMasker's performance drops as dataset size decreases. Next, ReMasker is mostly evaluated on datasets with at least 1,000 rows (Tab. 6 in \cite{du2023remasker}. CACTI is evaluated only on datasets with at least 3,539 rows and half having over 15,000 rows (Tab. A9 in \cite{gorla2025cacti}). In MissBench, however, we focus on small tables, resulting in a low-information setting. Notably, we also tried to test against DiffPuter \citep{zhang2024diffputer}, but found that it did not converge well on the smaller tables we tested on. We believe this was not an issue in their original paper because they only test on datasets with at least 9,522 rows and most tables in their tests have over 15,000 rows. TabImpute models, on the other hand, are pre-trained, which allows them to perform well with less information by leveraging their strong prior. These results highlight the difficulty of imputing missing data in smaller tables and suggest that different imputation methods may be complementary across data regimes.

\section{Conclusion \& Future Work}

In this paper, we present a flexible pre-training pipeline for imputing missing data in tabular settings. By combining an entry-wise featurization with a synthetic data generator whose missingness module is interchangeable, our pipeline produces zero-shot transformer-based imputers that can be tailored to any target missingness pattern, without modifying the model architecture or training algorithm. We show that specialized TabImpute models produced by the pipeline outperform existing pattern-specific methods, while the default TabImpute model trained on \mcar alone is robust across \numpatterns diverse missingness patterns on real-world OpenML data. We open-source our pipeline, model weights, and evaluation code, and we hope this will enable practitioners to train imputers tailored to their specific missingness settings without the need to design new algorithms.

While our pipeline performs well on small and medium-sized tables, the entry-wise featurization introduces an $mn$-row expansion, which limits scalability to very large tables. We plan to reduce this computational complexity and expand to larger tables. Recent work on tabular representation learning has shown possible paths forward: \citep{zeng2025tabflex} and \citep{qu2025tabicl} propose attention mechanisms that improve the scalability of tabular foundation models. We also plan to (i) incorporate more complex data-generating processes into the synthetic prior, (ii) enhance the pipeline to better handle categorical data by integrating discrete variables into pre-training, (iii) extend evaluation to causal-inference settings, which can be modeled as missing-data problems \citep{agarwal2023causal}, and (iv) explore how domain-specific knowledge about the missingness mechanism can be encoded in the synthetic generator to produce even more targeted imputers.

\newpage
\bibliographystyle{plain}
\bibliography{references}

\onecolumn
\appendix
\section{Appendix}
Here we present the rest of the missingness patterns we tested on, tables with further results, the methods we tested against, and the OpenML datasets we evaluated on.

\subsection{Full RMSE and $R^2$ results on MissBench}

\begin{figure}[!ht]
    \centering
    \includegraphics[width=\linewidth]{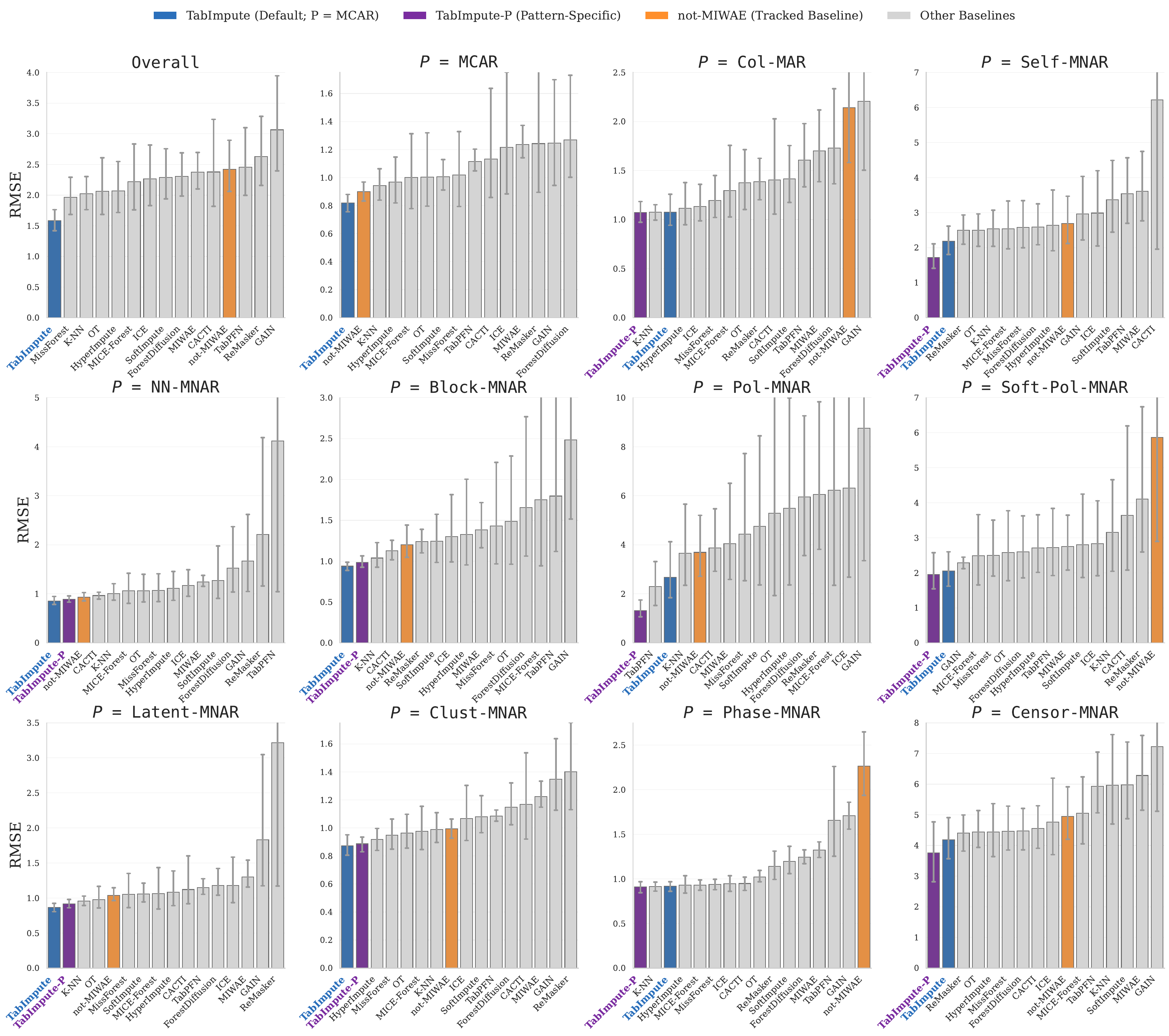}
    \caption{\textbf{RMSE for all patterns and overall.} Here, we include both the default TabImpute model (\mcar-trained) and the specialized models. Specialized TabImpute models are labeled with just the pattern name for readability. All TabImpute models are colored in blue, not-MIWAE in orange to show its variance in rank across patterns, and the other baselines are shown in gray.}
    \label{fig:all-bar-plot}
\end{figure}

\begin{table}[h]
\centering
\caption{RMSE $\pm$ 95\% confidence interval by missingness pattern}
\label{tab:normalized_negative_rmse_by_pattern_full}
\resizebox{\linewidth}{!}{\begin{tabular}{llccccc}
\toprule
& Pattern & TabImpute & MissForest & K-NN & OT & HyperImpute \\
\midrule
\textbf{\mcar} & \mcar & \textbf{0.819 ± 0.065} & 1.018 ± 0.291 & 0.943 ± 0.119 & 1.004 ± 0.279 & 0.969 ± 0.184 \\
\noalign{\vskip 1.5pt}
\hdashline
\noalign{\vskip 1.5pt}
\textbf{\mar} & \colmar & 1.076 ± 0.164 & 1.197 ± 0.229 & \textbf{1.075 ± 0.082} & 1.377 ± 0.340 & 1.117 ± 0.228 \\
\noalign{\vskip 1.5pt}
\hdashline
\noalign{\vskip 1.5pt}
\multirow{9}{*}{\textbf{\mnar}} & \mnarself & \textbf{2.184 ± 0.435} & 2.580 ± 0.716 & 2.539 ± 0.555 & 2.500 ± 0.499 & 2.636 ± 0.949 \\
 & \nnmar & \textbf{0.854 ± 0.083} & 1.070 ± 0.300 & 1.004 ± 0.194 & 1.065 ± 0.310 & 1.114 ± 0.310 \\
 & \marblockneural & \textbf{0.939 ± 0.051} & 1.430 ± 0.752 & 1.039 ± 0.161 & 1.489 ± 0.721 & 1.325 ± 0.656 \\
 & \polarmnar & 2.680 ± 1.318 & 4.433 ± 3.042 & 3.660 ± 1.755 & 5.290 ± 4.726 & 5.492 ± 4.183 \\
 & \softpolarmnar & \textbf{2.051 ± 0.515} & 2.498 ± 0.828 & 3.157 ± 1.428 & 2.584 ± 1.038 & 2.713 ± 0.901 \\
 & \latentmnar & \textbf{0.867 ± 0.060} & 1.052 ± 0.276 & 0.958 ± 0.070 & 0.975 ± 0.169 & 1.082 ± 0.257 \\
 & \clustermnar & \textbf{0.872 ± 0.075} & 0.947 ± 0.111 & 0.988 ± 0.120 & 0.965 ± 0.131 & 0.917 ± 0.082 \\
 & \twophasemnar & 0.917 ± 0.058 & 0.937 ± 0.061 & \textbf{0.915 ± 0.055} & 1.022 ± 0.067 & 0.931 ± 0.095 \\
 & \censormnar & \textbf{4.181 ± 0.721} & 4.465 ± 0.759 & 5.961 ± 1.531 & 4.442 ± 0.622 & 4.444 ± 0.981 \\
\midrule
 & Overall & \textbf{1.585 ± 0.174} & 1.966 ± 0.325 & 2.022 ± 0.285 & 2.065 ± 0.456 & 2.067 ± 0.423 \\
\bottomrule
\end{tabular}}
\resizebox{\linewidth}{!}{\begin{tabular}{llccccc}
\toprule
& Pattern & MICE-Forest & ICE & SoftImpute & ForestDiffusion & MIWAE \\
\midrule
\textbf{\mcar} & \mcar & 1.002 ± 0.309 & 1.215 ± 0.470 & 1.007 ± 0.116 & 1.269 ± 0.435 & 1.235 ± 0.125 \\
\noalign{\vskip 1.5pt}
\hdashline
\noalign{\vskip 1.5pt}
\textbf{\mar} & \colmar & 1.298 ± 0.389 & 1.135 ± 0.195 & 1.417 ± 0.317 & 1.730 ± 0.525 & 1.701 ± 0.383 \\
\noalign{\vskip 1.5pt}
\hdashline
\noalign{\vskip 1.5pt}
\multirow{9}{*}{\textbf{\mnar}} & \mnarself & 2.543 ± 0.706 & 2.988 ± 1.137 & 3.372 ± 1.208 & 2.593 ± 0.619 & 3.608 ± 0.985 \\
 & \nnmar & 1.061 ± 0.336 & 1.171 ± 0.287 & 1.270 ± 0.588 & 1.522 ± 0.715 & 1.246 ± 0.121 \\
 & \marblockneural & 1.751 ± 1.179 & 1.299 ± 0.503 & 1.243 ± 0.327 & 1.654 ± 1.082 & 1.384 ± 0.305 \\
 & \polarmnar & 6.231 ± 5.506 & 6.318 ± 5.183 & 4.750 ± 3.356 & 5.957 ± 3.279 & 4.044 ± 2.232 \\
 & \softpolarmnar & 2.492 ± 1.094 & 2.829 ± 1.110 & 2.804 ± 1.267 & 2.601 ± 0.936 & 2.753 ± 0.840 \\
 & \latentmnar & 1.062 ± 0.348 & 1.181 ± 0.361 & 1.056 ± 0.140 & 1.177 ± 0.213 & 1.301 ± 0.231 \\
 & \clustermnar & 0.977 ± 0.160 & 1.068 ± 0.200 & 1.079 ± 0.142 & 1.148 ± 0.155 & 1.224 ± 0.100 \\
 & \twophasemnar & 0.931 ± 0.057 & 0.947 ± 0.096 & 1.198 ± 0.159 & 1.245 ± 0.077 & 1.322 ± 0.092 \\
 & \censormnar & 5.051 ± 1.066 & 4.762 ± 1.323 & 5.978 ± 1.315 & 4.477 ± 0.766 & 6.284 ± 1.330 \\
\midrule
 & Overall & 2.217 ± 0.542 & 2.265 ± 0.519 & 2.289 ± 0.388 & 2.307 ± 0.363 & 2.373 ± 0.297 \\
\bottomrule
\end{tabular}}\\
\resizebox{\linewidth}{!}{\begin{tabular}{llccccc}
\toprule
& Pattern & CACTI & not-MIWAE & TabPFN & ReMasker & GAIN \\
\midrule
\textbf{\mcar} & \mcar & 1.132 ± 0.463 & 0.900 ± 0.070 & 1.115 ± 0.082 & 1.243 ± 0.480 & 1.246 ± 0.389 \\
\noalign{\vskip 1.5pt}
\hdashline
\noalign{\vskip 1.5pt}
\textbf{\mar} & \colmar & 1.405 ± 0.576 & 2.141 ± 0.664 & 1.608 ± 0.335 & 1.385 ± 0.213 & 2.207 ± 1.137 \\
\noalign{\vskip 1.5pt}
\hdashline
\noalign{\vskip 1.5pt}
\multirow{9}{*}{\textbf{\mnar}} & \mnarself & 6.219 ± 8.109 & 2.686 ± 0.706 & 3.536 ± 0.980 & 2.499 ± 0.463 & 2.965 ± 0.992 \\
 & \nnmar & 0.965 ± 0.074 & 0.933 ± 0.093 & 4.117 ± 6.103 & 2.212 ± 1.901 & 1.667 ± 0.863 \\
 & \marblockneural & 1.128 ± 0.130 & 1.199 ± 0.225 & 1.797 ± 1.246 & 1.241 ± 0.159 & 2.484 ± 1.780 \\
 & \polarmnar & 3.876 ± 1.379 & 3.692 ± 1.356 & \textbf{2.295 ± 0.989} & 6.059 ± 3.475 & 8.751 ± 7.595 \\
 & \softpolarmnar & 3.638 ± 2.191 & 5.862 ± 4.215 & 2.721 ± 0.971 & 4.107 ± 2.477 & 2.287 ± 0.182 \\
 & \latentmnar & 1.121 ± 0.339 & 1.038 ± 0.099 & 1.147 ± 0.119 & 3.218 ± 3.813 & 1.833 ± 1.203 \\
 & \clustermnar & 1.169 ± 0.341 & 0.995 ± 0.070 & 1.085 ± 0.042 & 1.402 ± 0.336 & 1.348 ± 0.273 \\
 & \twophasemnar & 0.948 ± 0.078 & 2.266 ± 0.375 & 1.656 ± 0.549 & 1.142 ± 0.158 & 1.711 ± 0.168 \\
 & \censormnar & 4.554 ± 0.771 & 4.946 ± 0.864 & 5.932 ± 1.067 & 4.403 ± 0.636 & 7.222 ± 3.128 \\
\midrule
 & Overall & 2.378 ± 0.769 & 2.424 ± 0.433 & 2.455 ± 0.591 & 2.628 ± 0.554 & 3.065 ± 0.791 \\
\bottomrule
\end{tabular}}
\end{table}

\begin{table}[h]
\centering
\caption{Mean column-wise $R^2$ $\pm$ 95\% confidence interval by missingness pattern}
\label{tab:r_squared_by_pattern}
\resizebox{\linewidth}{!}{
\begin{tabular}{llccccc}
\toprule
& Pattern & TabImpute & HyperImpute & ICE & MICE-Forest & MissForest \\
\midrule
\textbf{\mcar} & \mcar & 0.414 ± 0.070 & 0.409 ± 0.071 & 0.387 ± 0.071 & \textbf{0.421 ± 0.072} & 0.401 ± 0.072 \\
\noalign{\vskip 1.5pt}
\hdashline
\noalign{\vskip 1.5pt}
\textbf{\mar} & \colmar & \textbf{0.289 ± 0.070} & 0.285 ± 0.070 & 0.287 ± 0.068 & 0.252 ± 0.072 & 0.242 ± 0.065 \\
\noalign{\vskip 1.5pt}
\hdashline
\noalign{\vskip 1.5pt}
\multirow{9}{*}{\textbf{\mnar}} & \mnarself & \textbf{0.285 ± 0.079} & 0.234 ± 0.077 & 0.250 ± 0.075 & 0.243 ± 0.071 & 0.207 ± 0.060 \\
 & \nnmar & \textbf{0.382 ± 0.068} & 0.364 ± 0.068 & 0.348 ± 0.067 & 0.381 ± 0.067 & 0.351 ± 0.067 \\
 & \marblockneural & \textbf{0.256 ± 0.066} & 0.242 ± 0.069 & 0.229 ± 0.066 & 0.226 ± 0.059 & 0.221 ± 0.068 \\
 & \polarmnar & \textbf{0.190 ± 0.038} & 0.189 ± 0.041 & 0.177 ± 0.042 & 0.173 ± 0.035 & 0.156 ± 0.035 \\
 & \softpolarmnar & \textbf{0.148 ± 0.049} & 0.146 ± 0.049 & 0.137 ± 0.047 & 0.123 ± 0.040 & 0.131 ± 0.045 \\
 & \latentmnar & 0.338 ± 0.066 & 0.335 ± 0.069 & 0.331 ± 0.066 & \textbf{0.339 ± 0.067} & 0.322 ± 0.069 \\
 & \clustermnar & 0.342 ± 0.070 & 0.342 ± 0.075 & 0.325 ± 0.073 & \textbf{0.343 ± 0.072} & 0.325 ± 0.072 \\
 & \twophasemnar & 0.331 ± 0.075 & 0.333 ± 0.074 & \textbf{0.334 ± 0.073} & 0.307 ± 0.069 & 0.310 ± 0.068 \\
 & \censormnar & \textbf{0.143 ± 0.040} & 0.121 ± 0.038 & 0.125 ± 0.038 & 0.095 ± 0.020 & 0.097 ± 0.034 \\
\midrule
 & Overall & \textbf{0.283 ± 0.020} & 0.273 ± 0.021 & 0.266 ± 0.020 & 0.264 ± 0.020 & 0.251 ± 0.020 \\
\bottomrule
\end{tabular}
}
\quad
\resizebox{\linewidth}{!}{
\begin{tabular}{llccccc}
\toprule
& Pattern & OT & CACTI & not-MIWAE & K-NN & GAIN \\
\midrule
\textbf{\mcar} & \mcar & 0.401 ± 0.068 & 0.337 ± 0.074 & 0.369 ± 0.067 & 0.323 ± 0.072 & 0.318 ± 0.062 \\
\noalign{\vskip 1.5pt}
\hdashline
\noalign{\vskip 1.5pt}
\textbf{\mar} & \colmar & 0.221 ± 0.063 & 0.242 ± 0.060 & 0.210 ± 0.057 & 0.227 ± 0.065 & 0.156 ± 0.054 \\
\noalign{\vskip 1.5pt}
\hdashline
\noalign{\vskip 1.5pt}
\multirow{9}{*}{\textbf{\mnar}} & \mnarself & 0.211 ± 0.069 & 0.203 ± 0.065 & 0.240 ± 0.061 & 0.207 ± 0.067 & 0.224 ± 0.059 \\
 & \nnmar & 0.334 ± 0.066 & 0.263 ± 0.072 & 0.340 ± 0.069 & 0.278 ± 0.063 & 0.265 ± 0.053 \\
 & \marblockneural & 0.199 ± 0.059 & 0.179 ± 0.062 & 0.165 ± 0.052 & 0.186 ± 0.059 & 0.098 ± 0.033 \\
 & \polarmnar & 0.177 ± 0.036 & 0.150 ± 0.039 & 0.144 ± 0.032 & 0.146 ± 0.037 & 0.109 ± 0.026 \\
 & \softpolarmnar & 0.108 ± 0.041 & 0.114 ± 0.045 & 0.056 ± 0.032 & 0.075 ± 0.036 & 0.078 ± 0.028 \\
 & \latentmnar & 0.306 ± 0.062 & 0.253 ± 0.067 & 0.265 ± 0.060 & 0.246 ± 0.062 & 0.202 ± 0.047 \\
 & \clustermnar & 0.314 ± 0.067 & 0.280 ± 0.071 & 0.280 ± 0.068 & 0.254 ± 0.065 & 0.239 ± 0.058 \\
 & \twophasemnar & 0.260 ± 0.060 & 0.285 ± 0.073 & 0.180 ± 0.051 & 0.293 ± 0.063 & 0.198 ± 0.052 \\
 & \censormnar & 0.084 ± 0.032 & 0.103 ± 0.023 & 0.121 ± 0.030 & 0.095 ± 0.021 & 0.127 ± 0.040 \\
\midrule
 & Overall & 0.238 ± 0.019 & 0.219 ± 0.019 & 0.215 ± 0.018 & 0.212 ± 0.018 & 0.183 ± 0.015 \\
\bottomrule
\end{tabular}
}
\quad
\resizebox{\linewidth}{!}{
\begin{tabular}{llccccc}
\toprule
& Pattern & ReMasker & SoftImpute & ForestDiffusion & TabPFN & MIWAE \\
\midrule
\textbf{\mcar} & \mcar & 0.327 ± 0.072 & 0.265 ± 0.084 & 0.220 ± 0.058 & 0.070 ± 0.017 & 0.036 ± 0.010 \\
\noalign{\vskip 1.5pt}
\hdashline
\noalign{\vskip 1.5pt}
\textbf{\mar} & \colmar & 0.185 ± 0.064 & 0.167 ± 0.066 & 0.108 ± 0.041 & 0.081 ± 0.023 & 0.030 ± 0.006 \\
\noalign{\vskip 1.5pt}
\hdashline
\noalign{\vskip 1.5pt}
\multirow{9}{*}{\textbf{\mnar}} & \mnarself & 0.151 ± 0.053 & 0.166 ± 0.060 & 0.116 ± 0.049 & 0.106 ± 0.033 & 0.026 ± 0.007 \\
 & \nnmar & 0.196 ± 0.064 & 0.223 ± 0.086 & 0.177 ± 0.052 & 0.042 ± 0.011 & 0.025 ± 0.007 \\
 & \marblockneural & 0.149 ± 0.062 & 0.165 ± 0.070 & 0.106 ± 0.033 & 0.051 ± 0.016 & 0.023 ± 0.005 \\
 & \polarmnar & 0.119 ± 0.039 & 0.118 ± 0.043 & 0.091 ± 0.025 & 0.036 ± 0.008 & 0.029 ± 0.007 \\
 & \softpolarmnar & 0.046 ± 0.025 & 0.077 ± 0.042 & 0.031 ± 0.008 & 0.004 ± 0.003 & 0.014 ± 0.003 \\
 & \latentmnar & 0.177 ± 0.059 & 0.194 ± 0.079 & 0.152 ± 0.045 & 0.053 ± 0.015 & 0.028 ± 0.007 \\
 & \clustermnar & 0.212 ± 0.068 & 0.187 ± 0.076 & 0.181 ± 0.054 & 0.058 ± 0.018 & 0.033 ± 0.011 \\
 & \twophasemnar & 0.263 ± 0.077 & 0.190 ± 0.072 & 0.126 ± 0.045 & 0.086 ± 0.024 & 0.022 ± 0.005 \\
 & \censormnar & 0.080 ± 0.025 & 0.081 ± 0.042 & 0.071 ± 0.022 & 0.072 ± 0.036 & 0.067 ± 0.037 \\
\midrule
 & Overall & 0.173 ± 0.018 & 0.166 ± 0.020 & 0.125 ± 0.013 & 0.060 ± 0.007 & 0.030 ± 0.004 \\
\bottomrule
\end{tabular}
}
\end{table}

\begin{table}[p]
\centering
\caption{\centering RMSE $\pm$ 95\% Confidence Interval for Specialized TabImpute Models \\The top-2 are bolded for each missingness pattern.}
\label{tab:specialized_models_comparison}
\resizebox{\textwidth}{!}{
\begin{tabular}{llc:c:ccc}
\toprule
& & \textbf{\mcar} & \textbf{\mar} & \multicolumn{3}{c}{\textbf{\mnar}} \\
\midrule
& Trained on $\Rightarrow$ & \shortstack{\mcar} & \shortstack{\colmar} & \shortstack{\texttt{Self}} & \shortstack{\texttt{NN}} & \shortstack{\texttt{Block}} \\
\midrule
\textbf{\mcar} & \mcar & \textbf{0.819 $\pm$ 0.065} & \textbf{0.871 $\pm$ 0.061} & 1.135 $\pm$ 0.068 & 1.153 $\pm$ 0.194 & 0.878 $\pm$ 0.052 \\
\noalign{\vskip 1.5pt}
\hdashline
\noalign{\vskip 1.5pt}
\textbf{\mar} & \colmar & \textbf{1.076 $\pm$ 0.164} & \textbf{1.072 $\pm$ 0.115} & 1.770 $\pm$ 0.292 & 1.442 $\pm$ 0.247 & 1.214 $\pm$ 0.194 \\
\noalign{\vskip 1.5pt}
\hdashline
\noalign{\vskip 1.5pt}
\multirow{9}{*}{\textbf{\mnar}} & \mnarself & \textbf{2.184 $\pm$ 0.435} & 2.214 $\pm$ 0.440 & \textbf{1.716 $\pm$ 0.381} & 2.957 $\pm$ 0.603 & 2.341 $\pm$ 0.410 \\
 & \nnmar & \textbf{0.854 $\pm$ 0.083} & 0.925 $\pm$ 0.052 & 1.070 $\pm$ 0.047 & 0.888 $\pm$ 0.069 & 0.934 $\pm$ 0.101 \\
 & \marblockneural & \textbf{0.939 $\pm$ 0.051} & 1.017 $\pm$ 0.064 & 1.304 $\pm$ 0.118 & 1.303 $\pm$ 0.371 & \textbf{0.985 $\pm$ 0.070} \\
 & \polarmnar & 2.680 $\pm$ 1.318 & \textbf{1.767 $\pm$ 0.417} & 2.780 $\pm$ 0.907 & 2.961 $\pm$ 2.023 & 2.065 $\pm$ 0.890 \\
 & \softpolarmnar & \textbf{2.051 $\pm$ 0.515} & 2.159 $\pm$ 0.642 & 2.391 $\pm$ 0.804 & 2.458 $\pm$ 0.946 & 2.095 $\pm$ 0.477 \\
 & \latentmnar & \textbf{0.867 $\pm$ 0.060} & 0.940 $\pm$ 0.058 & 1.097 $\pm$ 0.057 & 1.097 $\pm$ 0.113 & 0.928 $\pm$ 0.066 \\
 & \clustermnar & \textbf{0.872 $\pm$ 0.075} & 0.922 $\pm$ 0.040 & 1.084 $\pm$ 0.046 & 1.086 $\pm$ 0.103 & 0.903 $\pm$ 0.049 \\
 & \twophasemnar & \textbf{0.917 $\pm$ 0.058} & 0.938 $\pm$ 0.060 & 1.282 $\pm$ 0.137 & 1.137 $\pm$ 0.061 & 0.974 $\pm$ 0.052 \\
 & \censormnar & 4.181 $\pm$ 0.721 & 4.220 $\pm$ 0.521 & \textbf{3.018 $\pm$ 0.475} & 5.425 $\pm$ 0.783 & 4.427 $\pm$ 0.637 \\
\midrule
 & Overall & \textbf{1.585 $\pm$ 0.174} & \textbf{1.549 $\pm$ 0.128} & 1.695 $\pm$ 0.138 & 1.991 $\pm$ 0.249 & 1.613 $\pm$ 0.147 \\
\bottomrule
\end{tabular}}\\
\resizebox{\textwidth}{!}{
\begin{tabular}{llccccc}
\toprule
& & \multicolumn{5}{c}{\textbf{\mnar}} \\
\midrule
& Trained on $\Rightarrow$ & \shortstack{\texttt{Pol}} & \shortstack{\texttt{Soft-Pol}} & \shortstack{\texttt{Latent}} & \shortstack{\texttt{Clust}} & \shortstack{\texttt{Phase}} \\
\midrule
\textbf{\mcar} & \mcar & 1.037 $\pm$ 0.011 & 1.653 $\pm$ 0.697 & 0.890 $\pm$ 0.064 & 0.874 $\pm$ 0.071 & 1.043 $\pm$ 0.022 \\
\noalign{\vskip 1.5pt}
\hdashline
\noalign{\vskip 1.5pt}
\textbf{\mar} & \colmar & 1.358 $\pm$ 0.171 & 2.471 $\pm$ 1.618 & 1.163 $\pm$ 0.140 & 1.082 $\pm$ 0.107 & 1.125 $\pm$ 0.141 \\
\noalign{\vskip 1.5pt}
\hdashline
\noalign{\vskip 1.5pt}
\multirow{9}{*}{\textbf{\mnar}} & \mnarself & 2.839 $\pm$ 0.493 & 5.117 $\pm$ 2.714 & 2.446 $\pm$ 0.479 & 2.330 $\pm$ 0.489 & 2.341 $\pm$ 0.451 \\
 & \nnmar & 1.039 $\pm$ 0.010 & 1.808 $\pm$ 0.881 & 0.908 $\pm$ 0.073 & \textbf{0.886 $\pm$ 0.068} & 1.042 $\pm$ 0.026 \\
 & \marblockneural & 1.071 $\pm$ 0.037 & 2.268 $\pm$ 1.692 & 1.007 $\pm$ 0.090 & 0.994 $\pm$ 0.108 & 1.056 $\pm$ 0.032 \\
 & \polarmnar & \textbf{1.316 $\pm$ 0.400} & 7.547 $\pm$ 6.923 & 2.096 $\pm$ 0.967 & 2.456 $\pm$ 1.311 & 3.479 $\pm$ 3.769 \\
 & \softpolarmnar & 2.207 $\pm$ 0.583 & \textbf{1.953 $\pm$ 0.527} & 2.207 $\pm$ 0.631 & 2.057 $\pm$ 0.457 & 2.717 $\pm$ 1.083 \\
 & \latentmnar & 1.043 $\pm$ 0.014 & 1.566 $\pm$ 0.673 & 0.915 $\pm$ 0.066 & \textbf{0.894 $\pm$ 0.066} & 1.038 $\pm$ 0.034 \\
 & \clustermnar & 1.032 $\pm$ 0.007 & 1.377 $\pm$ 0.288 & 0.895 $\pm$ 0.045 & \textbf{0.887 $\pm$ 0.053} & 1.027 $\pm$ 0.023 \\
 & \twophasemnar & 1.158 $\pm$ 0.066 & 1.352 $\pm$ 0.150 & 0.996 $\pm$ 0.062 & 0.934 $\pm$ 0.065 & \textbf{0.911 $\pm$ 0.065} \\
 & \censormnar & 5.066 $\pm$ 0.622 & 7.971 $\pm$ 2.217 & 4.638 $\pm$ 0.589 & 4.483 $\pm$ 0.642 & 5.247 $\pm$ 0.856 \\
\midrule
 & Overall & 1.742 $\pm$ 0.144 & 3.189 $\pm$ 0.752 & 1.651 $\pm$ 0.159 & 1.625 $\pm$ 0.173 & 1.911 $\pm$ 0.374 \\
\bottomrule
\end{tabular}}\\
\begin{tabular}{llc}
\toprule
& & \textbf{\mnar} \\
\midrule
& Trained on $\Rightarrow$ & \shortstack{\texttt{Censor}} \\
\midrule
\textbf{\mcar} & \mcar & 2.259 $\pm$ 0.465 \\
\noalign{\vskip 1.5pt}
\hdashline
\noalign{\vskip 1.5pt}
\textbf{\mar} & \colmar & 2.526 $\pm$ 0.668 \\
\noalign{\vskip 1.5pt}
\hdashline
\noalign{\vskip 1.5pt}
\multirow{9}{*}{\textbf{\mnar}} & \mnarself & 3.593 $\pm$ 1.726 \\
 & \nnmar & 2.014 $\pm$ 0.293 \\
 & \marblockneural & 2.288 $\pm$ 0.765 \\
 & \polarmnar & 8.324 $\pm$ 3.811 \\
 & \softpolarmnar & 2.465 $\pm$ 0.362 \\
 & \latentmnar & 2.083 $\pm$ 0.442 \\
 & \clustermnar & 1.921 $\pm$ 0.219 \\
 & \twophasemnar & 2.198 $\pm$ 0.184 \\
 & \censormnar & \textbf{3.758 $\pm$ 1.004} \\
\midrule
 & Overall & 3.039 $\pm$ 0.426 \\
\bottomrule
\end{tabular}

\end{table}

\begin{figure}[ht!]
    \centering
    \includegraphics[width=0.5\linewidth]{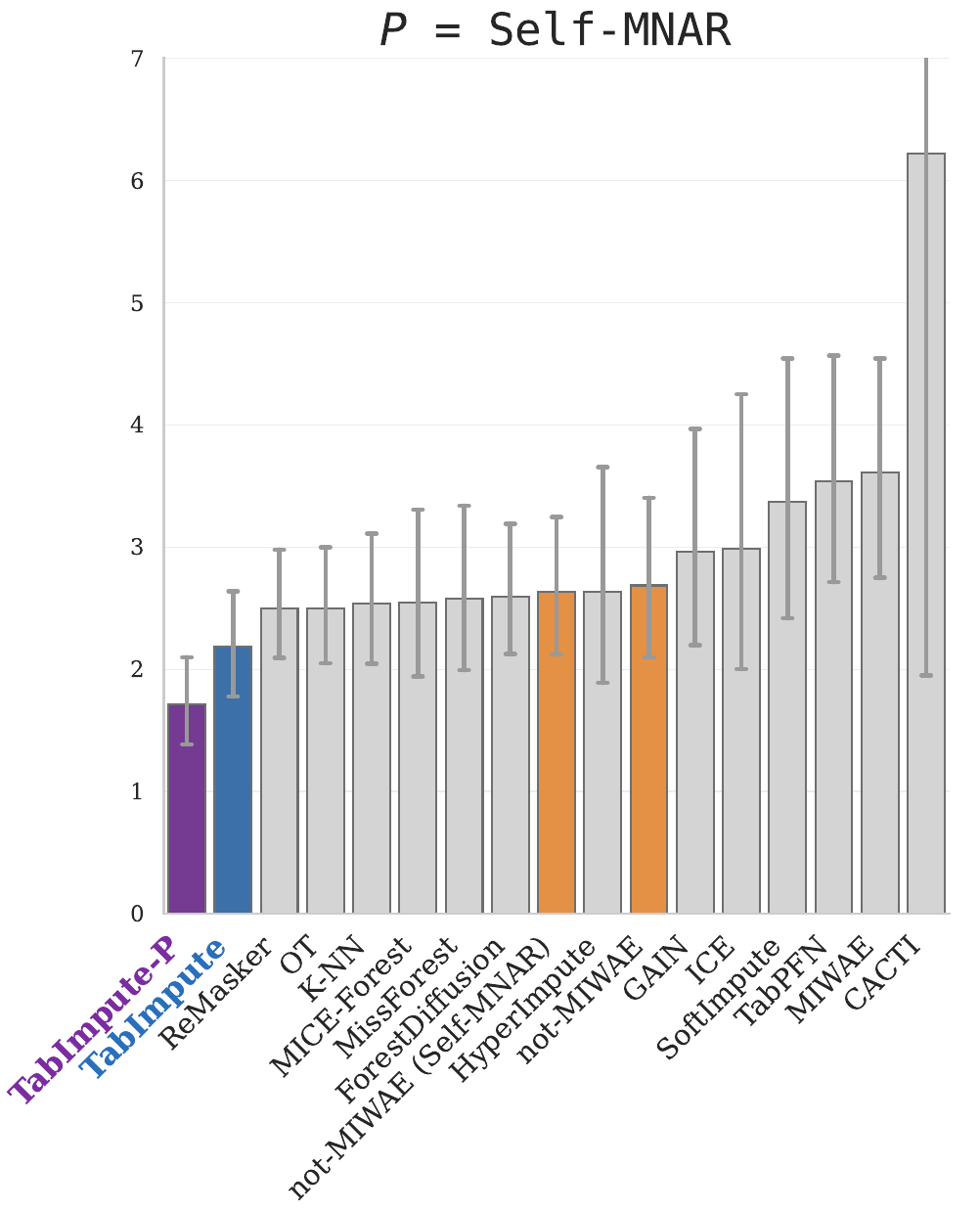}
    \caption{\textbf{RMSE including the specialized not-MIWAE designed for \mnarself.} Both the default TabImpute and TabImpute-\mnarself outperform both versions of not-MIWAE.}
    \label{fig:not-miwae}
\end{figure}

\clearpage

\subsection{Varying missingness level}\label{app:add-test}
\begin{figure}[!h]
    \centering
    \includegraphics[width=1.0\linewidth]{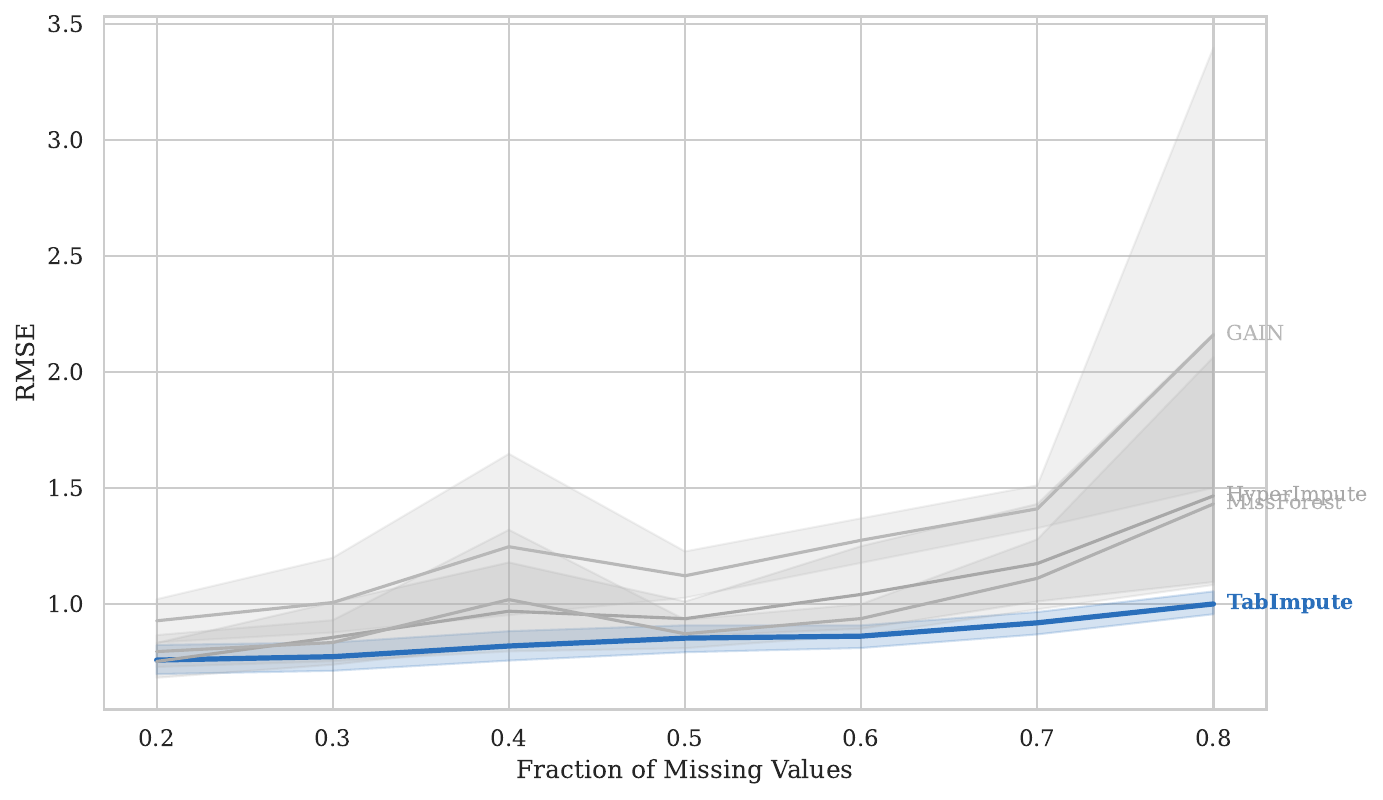}
    \caption{\textbf{RMSE vs. probability of missingness for \mcar.} \method  exhibits a remarkably robust performance as missingness increases and remain the best imputer across all missingness fractions.
    }
    \label{fig:mcar-prob}
\end{figure}

The probability of missingness can only be controlled precisely for \mcar. Thus, we show in \cref{fig:mcar-prob} the performance of the top methods as we increase the missingness level. We see that TabImpute's relative performance gets better as missingness increases.

\subsection{Details for \mnar Missingness Patterns}\label{app:mnar}
This section provides the mathematical and implementation details for each of the simulated \mnar missingness mechanisms that we implement and test. For each pattern, we define the mechanism by which the missingness mask $M$ is generated, where $M_{ij} = 1$ if the value $X_{ij}$ is observed and $M_{ij} = 0$ otherwise.

Recent work expands on Rubin's framework, arguing that missingness can exhibit rich multivariate structure that is not captured well by the standard \mcar/\mar/\mnar taxonomy. In particular, \citep{mitra2023learning} introduces structured missingness as an umbrella term for mechanisms in which missingness follows systematic patterns, and \citep{jackson2023completecharacterisationstructuredmissingness} provides a complementary characterization that accounts for dependencies among missingness indicators across variables. The diversity of missingness patterns encountered in practice underscores a core challenge: when a new pattern arises, practitioners must either rely on a general-purpose method that may underperform or wait for researchers to design a pattern-specific algorithm. For instance, \citep{agarwal2023causal} proposed a nearest neighbor-based matrix completion method specifically designed for a block-wise \mnar pattern, an approach that requires significant methodological effort and does not transfer to other patterns.

\subsubsection{Details for \nnmar}

\paragraph{Description:} This pattern simulates a scenario where the propensity $p_{ij}$ depends on the underlying matrix values $X^*$ in an arbitrary manner. We achieve a broad coverage of \mnar patterns by leveraging the expressiveness of neural networks.

\paragraph{Methodology:}
One general form of \mnar can be described as follows: for all $i$ and $j$, there exists some function $f_{ij}$ on the true (hence unobserved) matrix $X^*$ such that the propensity depends on $X^*$ as follows: $p_{ij}(X^*)=\P(M_{ij} = 1|X^*) = f_{ij}(X^*)$. By leveraging the expressiveness of neural networks, we propose a neural-net-based \mnar pattern generator (\nnmar) that is designed to approximate arbitrary propensities characterized by functions $f_{ij}$.

\paragraph{Implementation Details:}
 For fixed indices $i$ and $j$, \nnmar constructs the propensity $p_{ij}$ in a two-step procedure. First, we randomly collect a subset of values from the matrix $X^*$ and flatten them as a vector, say $X^*(i, j)$; we do this by first randomly generating a neighborhood $\mbf N_{ij}\subset [m]\times [n]$, then the entries in the neighborhood $X^*_{st}, (s, t) \in \mbf N_{ij}$ constitute the entries of the vector $X^*(i, j)$. Second, a neural-net function $g_{ij}:\R^{|\mbf N_{ij}|} \to [0, 1]$ is constructed by randomly initializing the number of layers, depth, weight, and bias. At each training step, the random neighborhood $\mbf N_{ij}$ and the random neural-net $g_{ij}$ collectively define the propensity $p_{ij} = g_{ij}(X^*(i, j))$ from which \mnar missingness patterns are generated by sampling $M_{ij}\sim \mrm{Bern}(p_{ij})$.

\subsubsection{\mnarself}
\paragraph{Description:} This pattern simulates a scenario where the probability of a value being missing is a direct function of the value itself. This pattern, commonly referred to as ``self-masked \mnar'', is widely utilized in the missing-data literature to model unbalanced class problems \citep{mohan2018handling, sportisse2020b}. A classic example occurs in income surveys, where individuals at the extremes of the income distribution, both very high and very low earners, are often less likely to disclose their income \citep{moore2000income}. Similar patterns appear in other domains, such as substance use reporting or medical contexts where patients with severe symptoms might be less able to complete follow-up assessments \citep{ibrahim2009missing}.

Within the structured missingness taxonomy proposed by \citep{jackson2023completecharacterisationstructuredmissingness}, this mechanism is classified as \textbf{MNAR-UP} (Unstructured, Probabilistic). It is ``unstructured'' because the missingness of an entry $(i, j)$ is conditionally independent of the missingness status of other entries, and ``probabilistic'' because it is determined by a probability function rather than a deterministic rule.
\paragraph{Methodology:}For each designated target column $j$, the probability of an entry $(i, j)$ being missing is determined by a logistic function of its value. The relationship is defined as $\P(M_{ij} = 0 | X^*_{ij}) = \sigma(\alpha \cdot X^*_{ij} + \beta_0)$ where $\sigma(z) = (1 + e^{-z})^{-1}$ is the sigmoid function.
\paragraph{Implementation Details:} A random coefficient $\alpha$ is chosen from the set $\{-2, -1, 1, 2\}$ to introduce variability in the direction and magnitude of the value's effect on its missingness probability. The bias term $\beta_0$ is calibrated to achieve a target missingness proportion.
\subsubsection{\censormnar}
\paragraph{Description:} 
This pattern models missingness arising from the physical limitations of measurement instruments, where values falling below a lower Limit of Detection or above an upper Limit of Quantification are not recorded. This results in left-censored or right-censored data, a common obstacle in environmental science, epidemiology, and biomedical research \citep{helsel2011statistics, chen2013bayesian}. In fields like metabolomics and proteomics, this type of missingness is explicitly recognized as left-censored MNAR \citep{karpievitch2009statistical, lazar2016accounting}, as the probability of being missing is directly determined by the concentration falling below the detection threshold. Failing to appropriately account for this, or using simplistic substitution methods like replacing with zero, can introduce substantial bias. Modeling \censormnar is therefore essential for evaluating imputation methods on datasets generated by instrumentation with inherent sensitivity constraints. 

\censormnar maps to the \textbf{MNAR-UD} (Unstructured, Deterministic) category in the structured missingness framework. It is ``Unstructured'' as the missingness of an entry is independent of other missingness indicators, and ``Deterministic'' because any value falling on the censored side of the threshold is missing with certainty.
\paragraph{Methodology:} For each column $j$, a censoring direction (left or right) is chosen with equal probability. A cutoff value is determined based on a specified quantile, $q_{\text{censor}}$, of the set of currently observed (non-missing) values in that column. Let $X^{*}_{:,j}$ denote the set of observed values in column $j$, i.e., $X^{*}_{:,j} = \{X_{ij} \mid M_{ij}=1\}$.
\begin{itemize}
    \item \textbf{Left-Censoring:} All values in column $j$ that are less than the $q_{\text{censor}}$-th quantile of the observed values in that same column are set to missing. The threshold is a single scalar value calculated from the column's observed data.
    $$ M_{ij} = 0 \quad \text{if} \quad X_{ij} < \text{quantile}(X^{*}_{:,j}, q_{\text{censor}}) $$

    \item \textbf{Right-Censoring:} All values in column $j$ that are greater than the $(1-q_{\text{censor}})$-th quantile of the observed values in that column are set to missing. The threshold is a single scalar value.
    $$ M_{ij} = 0 \quad \text{if} \quad X_{ij} > \text{quantile}(X^{*}_{:,j}, 1 - q_{\text{censor}}) $$
\end{itemize}
\paragraph{Implementation Details:} The choice between left- and right-censoring is made randomly for each column with a probability of 0.5 for each. We introduce a hyperparameter $q_{\text{censor}}$ for the censoring quantile that controls the fraction of data to be censored from either tail of the distribution. For our evaluation, we use $q_{\text{censor}}=0.25$.

\subsubsection{\polarmnar}
\paragraph{Description:} This pattern simulates data where values falling in the middle of a feature's distribution are preferentially removed, simulating survey non-response from individuals with moderate opinions. This is implemented by setting values between the $q$-th and $(1-q)$-th quantiles to missing. We consider both hard and soft variants. A soft version makes the observation probability proportional to the value's distance from the median. \polarmnar is recognized in the survey methodology literature as a form of voluntary response bias or self-selection bias \citep{bethlehem2010selection}. One obvious example can be found in online product reviews, which frequently exhibit a U-shaped or J-shaped distribution, heavily skewed toward the highest and lowest ratings while the middle ground remains sparse—sometimes referred to as the ``brag-and-moan'' bias \citep{hu2009overcoming}. Similarly, in political polling, nonresponse bias can be exacerbated if highly polarized individuals are more motivated to participate than moderates, potentially exaggerating measures of mass polarization \citep{cavari2023survey}.

This pattern maps to the ``unstructured'' MNAR categories of \cite{jackson2023completecharacterisationstructuredmissingness}. Specifically, our hard polarization variant, where values within a quantile range are missing with certainty, is an instance of \textbf{MNAR-UD} (Unstructured, Deterministic). Our soft polarization variant, where the missingness probability is a function of the value's distance from the median, is an example of \textbf{MNAR-UP} (Unstructured, Probabilistic).
\paragraph{Hard Polarization Methodology:} For each column $j$, values falling between two quantiles are deterministically masked. The quantiles are calculated using only the set of currently observed (non-'NaN') values in that column. Let $X^{*}_{:,j}$ denote this set of observed values, i.e., $X^{*}_{:,j} = \{X_{ij} \mid M_{ij}=1\}$. The lower and upper thresholds, $L_j$ and $H_j$, are defined as:
$$ L_j = \text{quantile}(X^{*}_{:,j}, q_{\text{thresh}}) $$
$$ H_j = \text{quantile}(X^{*}_{:,j}, 1 - q_{\text{thresh}}) $$
An entry $X_{ij}$ is then masked if its value falls between these two scalar thresholds:
$$ M_{ij} = 0 \quad \text{if} \quad L_j < X_{ij} < H_j $$
\paragraph{Soft Polarization Methodology:} The probability of a value being observed is made proportional to its normalized absolute distance from the column's median, $\mu_j$. This creates a softer, probabilistic version of the polarization effect. The missing probability is given by:
$$ \P(M_{ij}=0) = \epsilon + (1 - 2\epsilon) \frac{|X^*_{ij} - \mu_j|^\alpha}{\max_k(|X^*_{kj} - \mu_j|^\alpha)}$$
\paragraph{Implementation Details:} For the hard polarization pattern, we introduce a hyperparameter $\boldsymbol{q_{\text{thresh}}}$ for the threshold quantile that defines the central portion of the distribution to be masked. For the soft polarization pattern, we have an exponent parameter $\alpha$ that controls the intensity of the polarization effect. Higher values of $\alpha$ make the observation probability more sensitive to deviations from the median. In the soft version, we also have a baseline probability $\epsilon$ that ensures even values at the median have a non-zero chance of being observed. 

\subsubsection{\latentmnar}
\paragraph{Description:} This pattern generates a complex missingness structure where the probability of an entry being missing depends on unobserved (latent) characteristics of both its row and its column. This is most common in recommender systems and collaborative filtering, where the data (e.g., user ratings) are inherently \mnar \citep{marlin2009collaborative, steck2010training}. Users do not select items to rate uniformly; rather, their exposure to items and their decisions to provide a rating are strongly influenced by their underlying (latent) preferences, the values the system aims to predict \citep{schnabel2016recommendations}. This creates a selection bias where observed ratings are a biased sample of the full data. A standard approach in the literature assumes a low-rank structure for both the underlying data matrix and the missingness mechanism (propensity score matrix), implying they are governed by a shared set of latent factors \citep{sportisse2020aimputation, jin2022matrix}. Recent work in causal matrix completion explicitly models these latent confounders \citep{agarwal2023causal}.


\paragraph{Methodology:} The probability of an entry $(i, j)$ being observed is modeled using a low-rank bilinear model. The observation probability is given by the sigmoid of a dot product of latent factors plus bias terms:
$$ \P(M_{ij} = 1) = \sigma(u_i^T v_j + b_i + c_j) $$
where $u_i \in \mathbb{R}^k$ and $v_j \in \mathbb{R}^k$ are $k$-dimensional latent vectors for row $i$ and column $j$, and $b_i$ and $c_j$ are scalar biases for the row and column, respectively.
\paragraph{Implementation Details:} We specify the rank $k$ that defines the dimensionality of the latent space, sampled as an integer.  The elements of the latent factor matrices $U \in \mathbb{R}^{N \times k}$ and $V \in \mathbb{R}^{D \times k}$, and the bias vectors $b \in \mathbb{R}^N$ and $c \in \mathbb{R}^D$, are sampled independently from a standard normal distribution.

\subsubsection{\clustermnar}
\paragraph{Description:} This pattern induces missingness based on latent group-level characteristics. Rows and columns are first assigned to discrete clusters, and each cluster has a random effect that uniformly influences the observation probability of all its members. Such structures are prevalent in education (students clustered in schools), healthcare (patients clustered in hospitals), and cross-national surveys. In these settings, missing data patterns are often not independent across observations but rather clustered \citep{van2011multiple}. For example, in multi-center clinical trials or Cluster Randomized Trials, missingness can be heavily influenced by site-specific factors such as resource availability or staff training \citep{diaz2014missing}. Ignoring this hierarchical structure during imputation fails to account for the within-cluster correlation and can lead to biased estimates \citep{enders2016multilevel}. Because the missingness mechanism is related to these cluster-level effects, which may themselves be latent, the missingness is \mnar.

In the structured missingness taxonomy, we classify \clustermnar as \textbf{MNAR-UP} (Unstructured, Probabilistic) due to the probabilistic dependence on unobserved factors, without dependence on other missingness indicators.
\paragraph{Methodology:} The probability of an entry $(i, j)$ being observed is determined by an additive model of random effects corresponding to the cluster assignments of its row $i$ and column $j$. Denoting the row assignments by $C_R(i)$ and column assignments by $C_C(j)$, the observation probability is modeled as: 
$$ \P(M_{ij} = 1) = \sigma(g_{C_R(i)} + h_{C_C(j)} + \epsilon_{ij}) $$
where: 
\begin{itemize}
    \item $\sigma(z) = (1 + e^{-z})^{-1}$ is the sigmoid function.
    \item $g_k \sim \mathcal{N}(0, \tau_r^2)$ is the random effect for row cluster $k$.
    \item $h_l \sim \mathcal{N}(0, \tau_c^2)$ is the random effect for column cluster $l$.
    \item $\epsilon_{ij} \sim \mathcal{N}(0, \epsilon_{\text{std}}^2)$ is an entry-specific noise term.
\end{itemize}
\paragraph{Implementation Details:} For a matrix with $N$ and $D$ columns, row assignments $C_R(i)$ are drawn uniformly from $\{0, ..., K_R-1\}$ and column assignments $C_C(j)$ are drawn uniformly from $\{0, ..., K_C-1\}$, where $K_R$ and $K_C$ are the total number of row and column clusters, respectively. The number of row clusters $K_R$, the number of column clusters $K_C$, and the standard deviation of the random effects ($\tau_r$, $\tau_c$, $\epsilon_{\text{std}})$ are hyperparameters specified for the data generation process.

\subsubsection{\twophasemnar}
\paragraph{Description:} This pattern mimics multi-stage data collection, mirroring established methodologies such as ``two-phase sampling'' (or ``double sampling'') \citep{neyman1938contribution} and ``Planned Missing Data Designs'' \citep{graham2006planned}. These designs are frequently employed in large-scale surveys and epidemiological studies to manage costs and participant burden when certain variables are expensive or difficult to measure \citep{rhemtulla2016planned}. Another example includes market research where basic demographics are collected from all participants, but detailed purchasing behavior is only gathered from a subset, with missingness related to income level. In these designs, a full sample provides baseline information (``cheap'' features), and a subset is strategically selected for follow-up (``expensive'' features).
\paragraph{Methodology:} Let $\mathcal{F} = \{0, 1, ..., D-1\}$ be the set of all column indices in the data matrix $X^*$. This set is randomly partitioned into a ``cheap'' subset $\mathcal{C} \subset \mathcal{F}$ and an ``expensive'' subset $\mathcal{E} \subset \mathcal{F}$, such that $\mathcal{C} \cup \mathcal{E} = \mathcal{F}$ and $\mathcal{C} \cap \mathcal{E} = \emptyset$. By design, features in the cheap set $\mathcal{C}$ are always observed.

The decision to collect the expensive features for a given row $i$ is based on a logistic model applied to its cheap features. Let $X^*_{i, \mathcal{C}}$ denote the vector of values $\{X_{ij} \mid j \in \mathcal{C}\}$ for row $i$. A score is calculated for each row:
$$ s_i = \text{normalize}(X_{i, \mathcal{C}}^{*T} w) $$
where $w$ is a vector of random weights and the normalize function applies z-score normalization to the resulting scores across all rows.

The probability that all expensive features are observed for row $i$ is then given by:
$$ \mathbb{P}(M_{ij} = 1 \text{ for all } j \in \mathcal{E}) = \sigma(\alpha + \beta \cdot s_i) $$
If the expensive features for row $i$ are not observed (based on the probability above), then all of its values in the expensive columns are masked as missing, i.e., $M_{ij} = 0$ for all $j \in \mathcal{E}$.
\paragraph{Implementation Details:} A fraction of columns are randomly assigned to be ``cheap.'' The weight vector for the scoring model is sampled from a standard normal distribution, $w \sim \mathcal{N}(0, 1)$. Parameters $\alpha, \beta$ control the base rate and score-dependency of the observation probability. In our implementation, they are set to default values of $\alpha=0$ and $\beta=2.0$.

\subsection{Additional tables}\label{app:add-tbl}

\begin{table}[!ht]
\centering
\caption{Other imputation methods}
\label{tab:imputation_methods}
\resizebox{\textwidth}{!}{\begin{tabular}{lp{8cm}}
\toprule
Name & Description \\
\midrule
SoftImpute \mbox{\citep{hastie2015matrix}} & Iterative soft thresholding singular value decomposition based on a low-rank assumption on the data. \\
$k$-Nearest Neighbors \mbox{\citep{60c19788-1128-3b5f-9275-2d63cc155832}} & Imputes each entry using the mean of its row-wise nearest neighbors. We set $k=5$ for our tests.\\
HyperImpute \mbox{\citep{jarrett2022hyperimpute}} & Iterative imputation method optimizing over a suite of imputation methods. \\
Optimal transport method \mbox{\citep{muzellec2020missing}} & Uses optimal transport distances as a loss to impute missing values based on the principle that two randomly drawn batches from the same dataset should share similar data distributions. \\
MissForest \mbox{\citep{missforest2011}} & Repeatedly trains a random forest model for each variable on the observed values to predict and fill in missing entries until convergence. \\
ICE \mbox{\citep{ice2011}} & Imputation with iterative and chained equations of linear/logistic models for conditional expectations.\\
MICE \mbox{\citep{royston2011multiple}} & Handles missing data by iteratively imputing each incomplete variable using regression models that condition on all other variables. For speed, we use the Mice-Forest package for MICE (\url{https://github.com/AnotherSamWilson/miceforest}).\\
GAIN \mbox{\citep{gain2018}} & Adapts generative adversarial networks \citep{goodfellow2020generative} where the generator imputes missing values and the discriminator identifies which components are observed versus imputed. \\
MIWAE \mbox{\citep{mattei2019miwae}} & Learns a deep latent variable model and then performs importance sampling for imputation. \\
ForestDiffusion \mbox{\citep{jolicoeur2024generating}} & Trains a diffusion model using XGBoost directly on incomplete tabular data and then fills in missing values with an adapted inpainting algorithm. \\
TabPFN \mbox{\citep{hollmann2023tabpfn}} & The \texttt{tabpfn-extensions} package includes a part to impute missing entries in a table column-by-column using TabPFN. \\
ReMasker \mbox{\citep{du2023remasker}} & A transformer-based method using masked autoencoding.\\
CACTI \mbox{\citep{gorla2025cacti}} & A transformer-based method that leverages copy-masking and text-based contextual information from column names. We use the version without textual information (CMAE) to keep our tests fair. \\
not-MIWAE \mbox{\citep{ipsen2020not}} & not-MIWAE is a method based on variational autoencoders that learns from incomplete data using importance weighting and is specifically designed to handle Missing Not At Random missingness without requiring an explicit model of the missingness process. \\
\bottomrule
\end{tabular}}
\end{table}

\begin{table}[h]
\centering
\caption{\centering RMSE $\pm$ 95\% confidence interval by Missingness Pattern\\
This table shows the default linear factor TabImpute model and a TabImpute model trained under a nonlinear factor model and \mcar missingess.}
\label{tab:errors_with_nonlinear}
\begin{tabular}{llcc}
\toprule
& Pattern & TabImpute & TabImpute (Nonlinear) \\
\midrule
\textbf{\mcar} & \mcar & \textbf{0.819 ± 0.065} & 0.822 ± 0.061 \\
\noalign{\vskip 1.5pt}
\hdashline
\noalign{\vskip 1.5pt}
\textbf{\mar} & \colmar & 1.076 ± 0.164 & \textbf{1.062 ± 0.106} \\
\noalign{\vskip 1.5pt}
\hdashline
\noalign{\vskip 1.5pt}
\multirow{9}{*}{\textbf{\mnar}} & \mnarself & \textbf{2.184 ± 0.435} & 2.295 ± 0.502 \\
 & \nnmar & \textbf{0.854 ± 0.083} & 0.885 ± 0.098 \\
 & \marblockneural & \textbf{0.939 ± 0.051} & 0.993 ± 0.100 \\
 & \polarmnar & 2.680 ± 1.318 & \textbf{2.503 ± 0.732} \\
 & \softpolarmnar & \textbf{2.051 ± 0.515} & 2.289 ± 0.848 \\
 & \latentmnar & \textbf{0.867 ± 0.060} & 0.881 ± 0.065 \\
 & \clustermnar & \textbf{0.872 ± 0.075} & 0.874 ± 0.058 \\
 & \twophasemnar & \textbf{0.917 ± 0.058} & 0.921 ± 0.061 \\
 & \censormnar & 4.181 ± 0.721 & \textbf{4.024 ± 0.471} \\
\midrule
 & Overall & \textbf{1.585 ± 0.174} & 1.595 ± 0.148 \\
\bottomrule
\end{tabular}
\end{table}

\begin{table}[!htbp]
\centering
\caption{Synthetic Data Generation Parameters}
\resizebox{\textwidth}{!}{%
\begin{tabular}{|l|l|c|c|}
\hline
\textbf{Missingness Pattern} & \textbf{Parameter Name} & \textbf{Symbol} & \textbf{Value} \\
\hline
\multirow{1}{*}{\mcar} & Missing probability & $p$ & 0.4 \\
\hline
\multirow{1}{*}{\colmar} & Missing probability & $p$ & 0.4 \\
\hline
\multirow{5}{*}{\nnmar}
& Neighborhood size & $|N_{ij}|$ & Variable \\
& Network layers & $L$ & Random \\
& Network depth & $d$ & Random \\
& Weight initialization & $W$ & Random \\
& Bias initialization & $b$ & Random \\
\hline
\multirow{2}{*}{\mnarself}
& Coefficient set & $\alpha$ & $\{-2, -1, 1, 2\}$ \\
& Target missing proportion & $p_{missing}$ & Variable \\
\hline
\multirow{6}{*}{\marblockneural}
& Missing probability & $p$ & 0.4 \\
& Matrix size N & $N$ & 100 \\
& Matrix size T & $T$ & 50 \\
& Row blocks & $B_r$ & 10 \\
& Column blocks & $B_c$ & 10 \\
& Convolution type & - & mean \\
\hline
\multirow{1}{*}{\polarmnar} & Threshold quantile & $q_{thresh}$ & 0.25 \\
\hline
\multirow{2}{*}{\softpolarmnar}
& Polarization alpha & $\alpha$ & 2.5 \\
& Polarization epsilon & $\epsilon$ & 0.05 \\
\hline
\multirow{2}{*}{\latentmnar}
& Latent rank (low) & $k_{low}$ & 1 \\
& Latent rank (high) & $k_{high}$ & 5 \\
\hline
\multirow{2}{*}{\clustermnar}
& Number of row clusters & $K_R$ & 5 \\
& Number of column clusters & $K_C$ & 4 \\
\hline
\multirow{1}{*}{\twophasemnar} & Cheap feature fraction & $f_{cheap}$ & 0.4 \\
\hline
\multirow{1}{*}{\censormnar} & Censoring quantile & $q_{censor}$ & 0.25 \\
\hline
\end{tabular}%
}
\label{tab:comprehensive_missingness_hyperparams}
\end{table}

\begin{table}[!ht]
\centering
\caption{OpenML datasets. All data is under the CC-BY 4.0 license via OpenML. OpenML's code is under the 3-Clause BSD license.}
\label{tab:openml_datasets}
\resizebox{\textwidth}{!}{\begin{tabular}{p{4cm}lll}
\hline
\textbf{Dataset} & \textbf{Size} & \textbf{Domain} & \textbf{Description} \\
\hline
EgyptianSkulls & $150 \times 5$ & Anthropology & Cranial measurements over time in Egypt \\
humans\_numeric & $75 \times 15$ & Biology & Human body measurements \\
FacultySalaries & $50 \times 5$ & Education/Economics & University faculty salary data \\
SMSA & $59 \times 16$ & Demographics/Economics & U.S. metropolitan statistical area data \\
Student-Scores & $56 \times 13$ & Education & Student exam scores \\
analcatdata\_election2000 & $67 \times 15$ & Political science & 2000 U.S. presidential election results \\
analcatdata\_gviolence & $74 \times 9$ & Criminology & Gun violence statistics \\
analcatdata\_olympic2000 & $66 \times 12$ & Sports/Economics & Olympic results and country stats \\
baskball & $96 \times 5$ & Sports analytics & Basketball performance data \\
visualizing\_hamster & $73 \times 6$ & Education/Toy & Example dataset for teaching \\
witmer\_census\_1980 & $50 \times 5$ & Demographics & U.S. census microdata (1980) \\
MercuryinBass & $53 \times 10$ & Environmental chemistry & Mercury concentrations in fish \\
SolarPower & $204 \times 5$ & Energy/Engineering & Solar power output records \\
WineDataset & $178 \times 14$ & Chemistry/Oenology & Wine physicochemical properties \\
alcohol-qcm-sensor & $125 \times 15$ & Analytical chemistry & Alcohol detection sensor readings \\
benzo32 & $195 \times 33$ & Chemistry/Toxicology & Benzodiazepine compound data \\
machine\_cpu & $209 \times 7$ & Computer systems & Predicting CPU performance \\
pwLinear & $200 \times 11$ & Mathematics/Engineering & Piecewise linear regression benchmark \\
pyrim & $74 \times 28$ & Chemistry/Pharmacology & Pyrimethamine bioassay compounds \\
slump & $103 \times 10$ & Civil engineering & Concrete slump test properties \\
ICU & $200 \times 20$ & Medicine & Intensive care patient data \\
appendicitis\_test & $106 \times 8$ & Medicine & Appendicitis diagnosis \\
appendicitis\_test\_edsa & $106 \times 8$ & Medicine & Educational appendicitis dataset \\
breast-cancer-coimbra & $116 \times 10$ & Medicine & Breast cancer diagnosis data \\
Rainfall-in-Kerala-1901-2017 & $117 \times 18$ & Climate science & Rainfall time series in Kerala \\
pollution & $60 \times 16$ & Environmental science & Air pollution measurements \\
treepipit & $86 \times 10$ & Ecology & Bird habitat distribution \\
autoPrice & $159 \times 16$ & Business/Economics & Automobile pricing dataset \\
dataset\_analcatdata\_creditscore & $100 \times 7$ & Finance & Credit scoring dataset \\
Swiss-banknote-conterfeit-detection & $200 \times 7$ & Finance/Fraud & Banknote authenticity classification \\
Glass-Classification & $214 \times 10$ & Forensics/Materials & Glass chemical composition (forensics) \\
chatfield\_4 & $235 \times 13$ & Statistics/Time series & Textbook time series data (Chatfield) \\
chscase\_vine1 & $52 \times 10$ & Agriculture/Statistics & Vine growth study \\
edm & $154 \times 18$ & Education & Student learning performance \\
metafeatures & $75 \times 32$ & Meta-learning & Dataset-level features \\
rabe\_131 & $50 \times 6$ & Chemistry/Benchmark & Spectroscopy regression dataset \\
rabe\_148 & $66 \times 6$ & Chemistry/Benchmark & Spectroscopy regression dataset \\
rabe\_265 & $51 \times 7$ & Chemistry/Benchmark & Spectroscopy regression dataset \\
sleuth\_case1201 & $50 \times 7$ & Statistics/Education & Applied regression textbook data \\
sleuth\_ex1605 & $62 \times 6$ & Statistics/Education & Applied regression textbook data \\
wisconsin & $194 \times 33$ & Medicine & Wisconsin breast cancer dataset \\
\hline
\end{tabular}}
\end{table}


\end{document}